\documentclass[acmlarge,9pt,prologue,table,xcdraw]{acmart}

\AtBeginDocument{%
  }

\setcopyright{acmlicensed}
\copyrightyear{2025}
\acmYear{2025}
\acmDOI{10.1145/3734869}

\usepackage{multirow}
\usepackage{multicol}
\usepackage{xspace}
\usepackage{subcaption}
\usepackage{hyperref}
\usepackage{amsmath}
\usepackage{graphicx}
\usepackage{xcolor}

\newcommand{\name}{\textit{GraPhy}\xspace}

\begin{document}
\def\thefootnote{\bf*}\footnotetext{Equal contribution.}
\def\thefootnote{\arabic{footnote}}

\title{Graph-Based Physics-Guided Urban PM2.5 Air Quality Imputation with Constrained Monitoring Data}

\author{Shangjie Du\textsuperscript{\bf*}}
\affiliation{
  \institution{University of California, Merced}
  \country{USA}}
\email{sdu14@ucmerced.edu}

\author{Hui Wei\textsuperscript{\bf*}}
\affiliation{
  \institution{University of California, Merced}
  \country{USA}}
\email{huiwei2@ucmerced.edu}

\author{Dong Yoon Lee}
\affiliation{
  \institution{University of California, Merced}
  \country{USA}}
\email{dlee267@ucmerced.edu}

\author{Zhizhang Hu}
\affiliation{
  \institution{University of California, Merced}
  \country{USA}}
\email{zhu42@ucmerced.edu}

\author{Shijia Pan}
\affiliation{
  \institution{University of California, Merced}
  \country{USA}}
\email{span24@ucmerced.edu}

\renewcommand{\shortauthors}{Du and Wei et al.}
\begin{abstract}
This work introduces \name, a graph-based, physics-guided learning framework for high-resolution and accurate air quality modeling in urban areas with limited monitoring data.
Fine-grained air quality monitoring information is essential for reducing public exposure to pollutants. However, monitoring networks are often sparse in socioeconomically disadvantaged regions, limiting the accuracy and resolution of air quality modeling. To address this, we propose a physics-guided graph neural network architecture called \name with layers and edge features designed specifically for low-resolution monitoring data. Experiments using data from California's socioeconomically disadvantaged San Joaquin Valley show that \name achieves the overall best performance evaluated by mean squared error (MSE), mean absolute error (MAE), and R-square value ($R^2$), improving the performance by 9\%-56\% compared to various baseline models. Moreover, \name consistently outperforms baselines across different spatial heterogeneity levels, demonstrating the effectiveness of our model design. \footnote{This work is an extension to the paper “GraPhy: Graph-Based Physics-Guided Urban Air Quality Modeling for Monitoring-Constrained Regions” published in the Proceedings of ACM International Conference on Systems for Energy-Efficient Built
Environments (BuildSys) 2024.}
\end{abstract}



\settopmatter{printfolios=true}
\pagenumbering{gobble}
\maketitle

\section{Introduction} \label{sec:introduction}
Urban air quality modeling has become a major focus of public health  \cite{kelly2015air, hu2023enhancing} and government decision-making \cite{NYTimes2024}.
According to the American Lung Association, 130 million Americans live in areas with unhealthy air pollution levels \cite{NBCNews2024}.
The United States Environmental Protection Agency (EPA) reports a strong correlation between air pollution and increased childhood asthma rates \cite{EPA2024}.
Therefore, accurate and fine-grained monitoring is essential for reducing pollutant exposure and its impact on public health. 

\begin{figure}
\includegraphics[width=\linewidth]{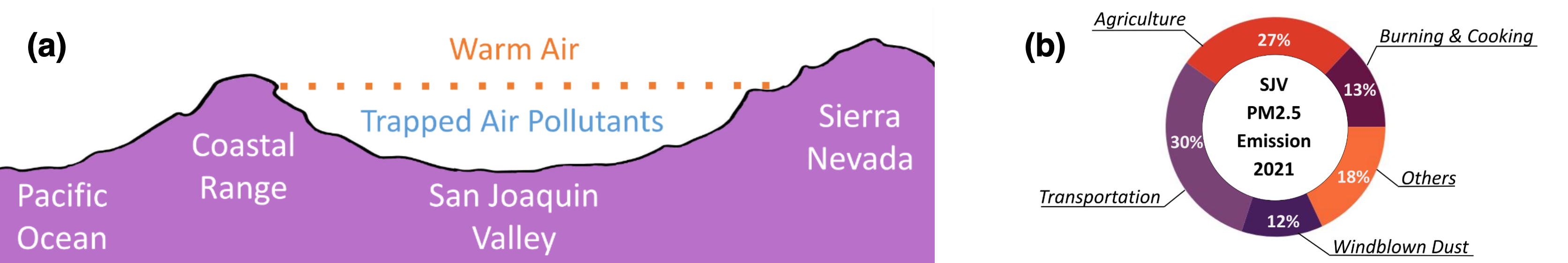}
    \caption{San Joaquin Valley (SJV)' challenges regarding air pollutant exposure: (a) The basin terrain of the SJV traps air pollutants easily; (b) The pollutant sources of PM2.5 in SJV (2021) show high variation \cite{SanJoaquinValleyAirPollutionControlDistrict2021}, which may cause the higher spatial heterogeneity in air pollutant measurements.}
    \label{fig:basin and sources}
\end{figure}

\begin{figure}
\includegraphics[width=\linewidth]{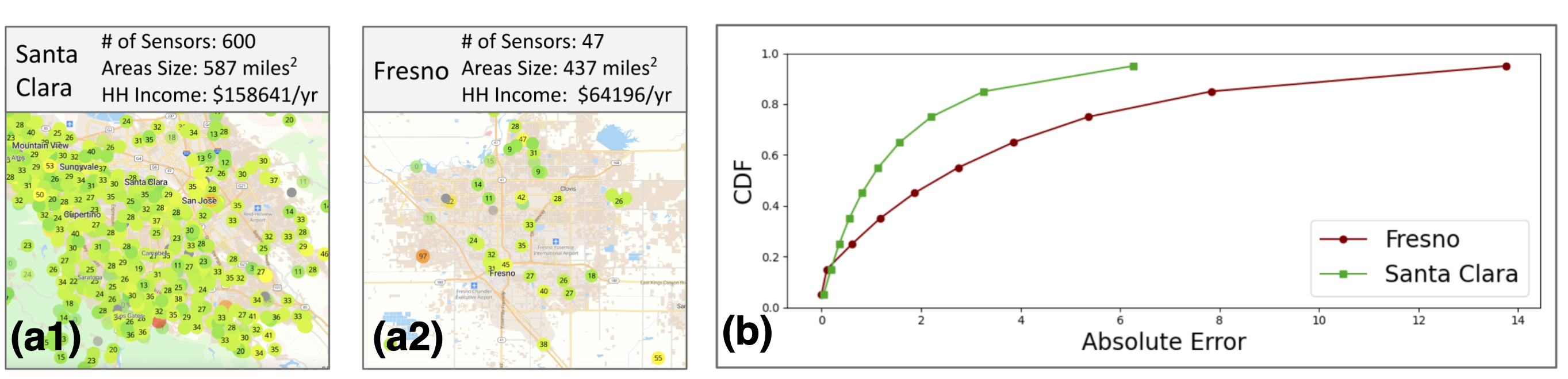}
    \vspace{-2ex}
    \caption{Purple Air sensor network deployment in Santa Clara and Fresno. (a) Santa Clara has a 10$\times$ deployment density and a 2.5$\times$ median household income compared to Fresno. (b) CDF of absolute errors using distance-based model \cite{shepard1968two} in Santa Clara and Fresno. Each point represents the CDF of absolute errors at 10\% percentiles across all interpolated data. The model for Fresno shows larger absolute errors at each percentile compared to Santa Clara, indicating that higher sensor density leads to more accurate predictions.}
    \vspace{-4ex}
    \label{fig:density and cdf}
\end{figure}

However, fine-grained monitoring requires a densely deployed air quality sensor system, which is infeasible for socioeconomically disadvantaged regions with constrained monitoring resources.
For instance, California's San Joaquin Valley is a region that is facing multi-fold challenges:
(1) Its unique geographical characteristics, as shown in Figure \ref{fig:basin and sources}(a), trap the polluted air in the valley, making it difficult to disperse and escape.
(2) Regional land uses for agriculture, oil extraction, and transportation result in significant emissions of pollutants \cite{SanJoaquinValleyAirPollutionControlDistrict2021}, as shown in Figure \ref{fig:basin and sources}(b).
(3) As a historically socioeconomically disadvantaged region, SJV has limited regulatory air quality monitoring stations and self-purchased air quality sensors.
For example, Figure \ref{fig:density and cdf}(a) compares the deployment density of the Purple Air Sensor Network\footnote{The US nationwide self-purchased air quality monitoring sensor network: map.purpleair.com} in Bay Area's Santa Clara and SJV's Fresno.
The deployment density in Fresno is only 1/10 of that in Santa Clara, which poses significant challenges in geospatial modeling \cite{hu2023enhancing}.
Figure \ref{fig:density and cdf}(b) shows a distance-based interpolation \cite{shepard1968two} algorithm's performance on data from these two cities, where the higher sensor density results in more accurate predictions in terms of absolute errors at all percentiles.
This result indicates the challenge of realizing fine-grained air quality monitoring in monitoring-constrained regions.

This work focuses on fine-grained urban air quality interpolation for monitoring-constrained regions.
The air quality spatial mapping task aims to predict the air quality at unmonitored locations and produce an air quality map for the area. Despite extensive research in recent years \cite{cheng2014aircloud, zheng2013u, hsieh2015inferring, patel2022accurate, chen2020adaptive, janssen2008spatial, martin2008satellite}, developing an accurate, fine-grained model for regions with \emph{extremely limited monitoring data} still faces following challenges:

\noindent \textbf{(C1) Sparsity of the air quality data increases modeling difficulty}.
Lower sensor density results in lower model accuracy (illustrated by Figure \ref{fig:density and cdf}).
Recent work on mobile sensing (e.g., on taxis) provides new solutions to address this issue \cite{liu2024mobiair, chen2020adaptive}. 
Despite its effectiveness in modern urban cities, it does not apply to suburban and rural regions with limited public transit or ride-sharing services. 

\noindent \textbf{(C2) Lack of fine-grained meteorology data impacts existing models' accuracy}.
Prior works leverage fine-grained (sub-kilometer level) meteorology information such as temperature, humidity, and wind data to improve air quality modeling accuracy \cite{zheng2013u, chen2020adaptive}.
However, the required fine-grained meteorology data is not guaranteed to be available for the studied area. Especially for monitoring-constrained areas, the entire area usually shares one value with no granularities.

\begin{figure}[htbp]
    \centering
    
    \begin{subfigure}[b]{0.40\textwidth}
        \centering
        \includegraphics[width=\linewidth]{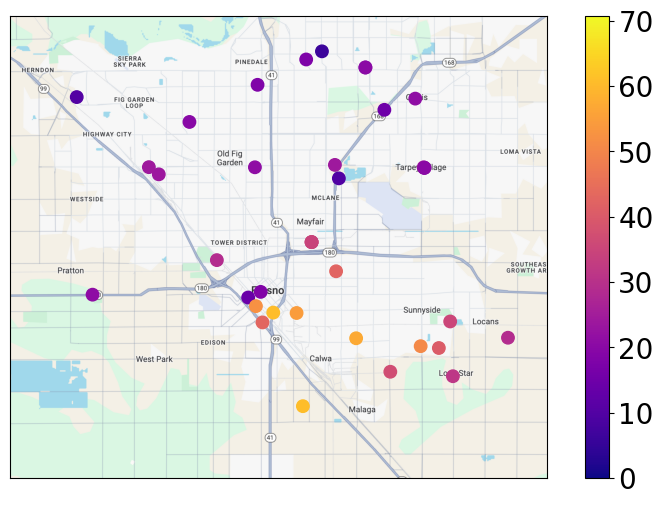} 
        \caption{High spatial heterogeneity}
        \label{fig:high_sh_example}
    \end{subfigure}%
    \hspace{0.5in}
    \begin{subfigure}[b]{0.40\textwidth}
        \centering
        \includegraphics[width=\linewidth]{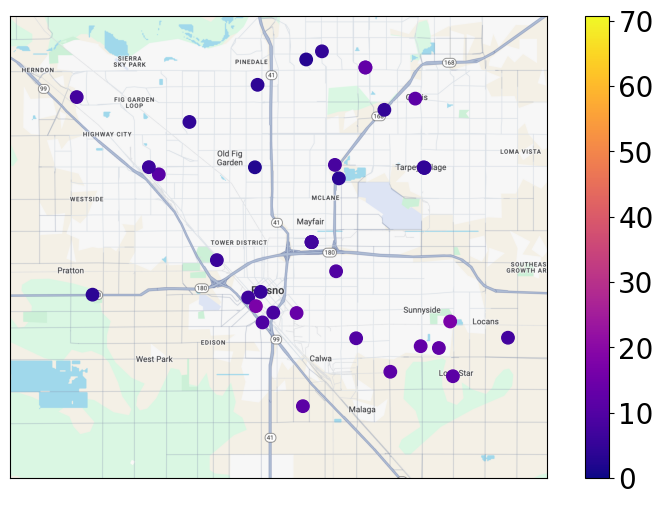} 
        \caption{Low spatial heterogeneity}
        \label{fig:low_sh_example}
    \end{subfigure}%
    \caption{Examples of high and low spatial heterogeneity (SH), where each point represents the PM2.5 concentration at a sensing location. Brighter colors indicate higher PM2.5 levels. (a) High spatial heterogeneity (SH = 206.77, calculated using Equation \ref{eq:sh} in Section \ref{sec:acc_sh}), showing significant variation in PM2.5 across the region at the same time. (b) Low spatial heterogeneity (SH = 13.20), indicating minimal variation in PM2.5 values across the region.}
    \label{fig:sh_examples}
\end{figure}

\noindent \textbf{(C3) Degradation of model performance under high spatial heterogeneity.}
The air quality spatial heterogeneity refers to the variation in air quality levels across different geographical locations. Figure \ref{fig:sh_examples} presents examples of high and low spatial heterogeneity.
Existing interpolation models tend to degrade under high spatial heterogeneity \cite{shepard1968two, wackernagel2003ordinary}. In the same deployment, higher spatial heterogeneity in measurements indicates a less uniform pollutant distribution, suggesting local sources or complex physical factors influencing air quality. This variability poses a significant challenge for accurate air quality modeling and prediction.

To address these challenges and improve air quality estimation in monitoring-constrained regions, we introduce \name, a physics-guided graph neural network architecture.
To allow more efficient spatial modeling with limited spatial measurements (\textbf{C1}), we adopt Graph Neural Networks (GNN) as the model architecture due to its flexibility in modeling both the individual sensor's measurements and the complex relationship between sensors.
To compensate for limited meteorology data (\textbf{C2}), we design customized wind-based edge features and a GNN model with residual message-passing as well as dual node-edge learners to make \name adaptive to coarse-grained meteorology data in monitoring-constrained regions.
To enhance the model accuracy under high spatial heterogeneity (\textbf{C3}), we leverage the Convection-Diffusion Equation \cite{schneider1992introduction} to incorporate physics domain knowledge, including learnable diffusion scaling, to guide our model's learning process. Also, we design a dynamic Softmax fusion method to adaptively fuse the information according to the input of air quality measurements.
We summarize our contributions as follows:
\begin{itemize}
    \item We introduce \name, a GNN-based architecture that integrates physics domain knowledge to estimate air quality in regions with limited monitoring resources.
    \item We specifically design message-passing processes for GNN modules to incorporate coarse-grained meteorology information, allowing it to adapt to regions with varying monitoring resource levels. We also introduce a learnable fusion mechanism to model dynamic ratios of three pollution propagation contributions, making the model robust against high spatial heterogeneity.
    \item Our evaluation on real-world air quality data from a monitoring-constrained region shows that \name outperforms all baseline models, achieving the overall best results in mean squared error (MSE), mean absolute error (MAE), and R-square ($R^2$), with performance gains of \textbf{9\%–56\%}. Additionally, \name consistently outperforms all baselines across different levels of spatial heterogeneity, demonstrating the effectiveness of our model design.
\end{itemize}

The rest of the paper is organized as follows: Section \ref{sec:background} covers the preliminaries on graph neural networks and convection-diffusion equations which our \name model builds on, as well as how graph convolution networks relate to the diffusion process. Section \ref{sec:methodology} introduces \name, detailing its design adaptations for constrained monitoring data. Section \ref{sec:eval} presents our evaluation and analysis of \name and various baseline models. Related work is discussed in Section \ref{sec:related}, followed by future directions in Section \ref{sec:discussion} and the conclusion in Section \ref{sec:conclusion}.

\section{Preliminaries}\label{sec:background}

This section introduces the key concepts of Graph Neural Networks, the core framework of our model, and the convection-diffusion equation, which inspires our model design.

\subsection{Graph Neural Networks (GNNs)} \label{subsec:GNN}
Graph Neural Networks are designed to model graph-structured data \cite{gori2005new, scarselli2008graph}, where the data are naturally represented as graphs of nodes and edges to capture the complex relationship.
A graph $G$ is defined as a pair $(V,E)$, where $V$ is a set of nodes and $E$ is a set of edges.
Let $v_i \in V$ denote a node and $e_{i,j}$ denote an edge between $v_i$ and $v_j$.
Nodes and edges in a graph are described by their features $x_v \in \mathbb{R}^d$ and $x_e \in \mathbb{R}^c$ ($d$ and $c$ are feature dimensions for a node and a edge, respectively) \cite{wu2020comprehensive}.
Similar to deep neural networks, GNNs are also constructed by stacking GNN layers.
When training the GNN, nodes and edges exchange information by message-passing (forward process), and the backpropagation of error will generate the gradient and update the learnable parameters that model the information flow between edges and nodes \cite{kipf2016semi}.
During the inference stage, GNN takes the input of nodes and edges' features from test samples and transforms them into the final prediction.

\vspace{1ex}
\noindent\textbf{Message-Passing Mechanism}
The message-passing mechanism \cite{gilmer2017neural} forms the foundation of Graph Neural Networks (GNNs), enabling information to flow between nodes and edges throughout the graph structure.
This determines the GNN layer's forward process.
The general form of the $k$th layer is described in Equation \ref{eq:messagepassing} below.
\begin{equation} \label{eq:messagepassing}
x_i^{(k)} = \gamma^{(k)} \left( x_i^{(k-1)}, \bigoplus_{j \in \mathcal{N}(i)} \, \phi^{(k)}\left(x_i^{(k-1)}, x_j^{(k-1)},e_{j,i}\right) \right)
\end{equation}
where $x_i^{(k-1)}$ is the node $i$'s features at the $(k-1)$th layer, $e_{j,i}$ is the edge from node $j$ to node $i$, and $\mathcal{N}(i)$ is the neighbors of node $i$.
The mechanism has three key functions: message $\phi^{(k)}$, aggregate $\bigoplus_{j \in \mathcal{N}(i)}$, and update $\gamma^{(k)}$.
For node $v_i$ and edge $e_{j, i}$, the message function $\phi^{(k)}$ takes node $v_i$'s, node $v_j$'s, and edge $e_{j,i}$'s features as inputs, and outputs the message propagated from node $v_j$ to node $v_i$.
The aggregate function $\bigoplus_{j \in \mathcal{N}(i)}$ combines all the propagated messages from neighboring nodes and outputs an aggregated message.
The update function $\gamma^{(k)}$ combines this aggregated message with $x_i^{(k-1)}$ and outputs $x_i^{(k)}$. 
This message-passing process is repeated by stacking multiple GNN layers to model more complex dependencies between nodes and edges.

\subsection{Convection-Diffusion Equation}
The convection-diffusion equation describes the combined effects of convection and diffusion on a physical quantity, such as particles and heat, in a given environment \cite{schneider1992introduction}. It has been widely used in the numerical modeling of urban air pollutant distribution \cite{randerson1970numerical, egan1972numerical, shir1974generalized}.
The general form of the convection-diffusion equation is:
\begin{equation} \label{eq:CD-Eq}
    \frac{\partial x}{\partial t} = d \nabla \cdot \nabla x - v \cdot \nabla x + R(s, t)
\end{equation}
where $x(s, t)$ is the scalar field of interest (e.g., PM2.5 at location $s$ and time $t$).
$v$ is the velocity vector field (e.g., wind field), representing the convection effect.
$d$ is the diffusion coefficient and $R(s, t)$ is the source or sink of the interested quantity.
Since the convection-diffusion equation incorporates two key transport mechanisms in air pollution modeling, it has been used with data-driven approaches to enhance spatiotemporal modeling in urban monitoring \cite{liu2022statistical}.

When it is used to model the PM2.5 concentration $x(s,t)$ at location $s$ and time $t$, the left side of Equation \ref{eq:CD-Eq} describes how the PM2.5 concentration changes with time $t$.
On the right side, the first term $d \nabla^2 x$ is the divergence of the PM2.5 concentration multiplied by the diffusion coefficient $d$, which describes the \emph{diffusion process of particles moving from high-concentration regions to low-concentration regions}.
The second term, $-v \cdot \nabla x$, describes \emph{how the PM2.5 changes due to the effect of the wind field.} 
It is determined by the field $v$ and the PM2.5 spatial gradient $\nabla x$.
The third term $R(s,t)$ accounts for the internal generation or consumption of PM2.5, which refers to \emph{the local source or sink of the air pollutant}.

\subsection{Diffusion on Graph Data with Graph Convolution Networks} \label{subsec:diffusion on graph}
The diffusion process introduced above can also be defined on a graph.
Graph diffusion is a process that spreads information across a graph structure, simulating how particles diffuse in physical systems \cite{bronstein2017geometric}.
This information flow can be implemented through the graph convolution operation, which is the matrix multiplication of convolution kernel $K$ and nodes' features $X$: $\text{conv}(X) = KX$.
This convolution enables each node to spread and exchange information with its neighbors.
On top of graph convolution operations, Graph Convolution Networks (GCNs) \cite{kipf2016semi} resemble the graph convolution with neural networks by adding trainable weights $W$ and nonlinear activation $\sigma$:
$$\text{GCN}(X) = \sigma(KXW)$$
Combining graph diffusion and GCNs \cite{hettige2024airphynet}, the trainable weights and the activation function of the neural networks enable a learnable and non-linear diffusion process on graphs, allowing the model to adaptively fit the diffusion patterns for a given data distribution.
The detailed design of our Diffusion GCN will be introduced in Section \ref{subsec:diffusion}.

\begin{figure*}[ht!]
    \centering
    \includegraphics[width=0.95\textwidth]{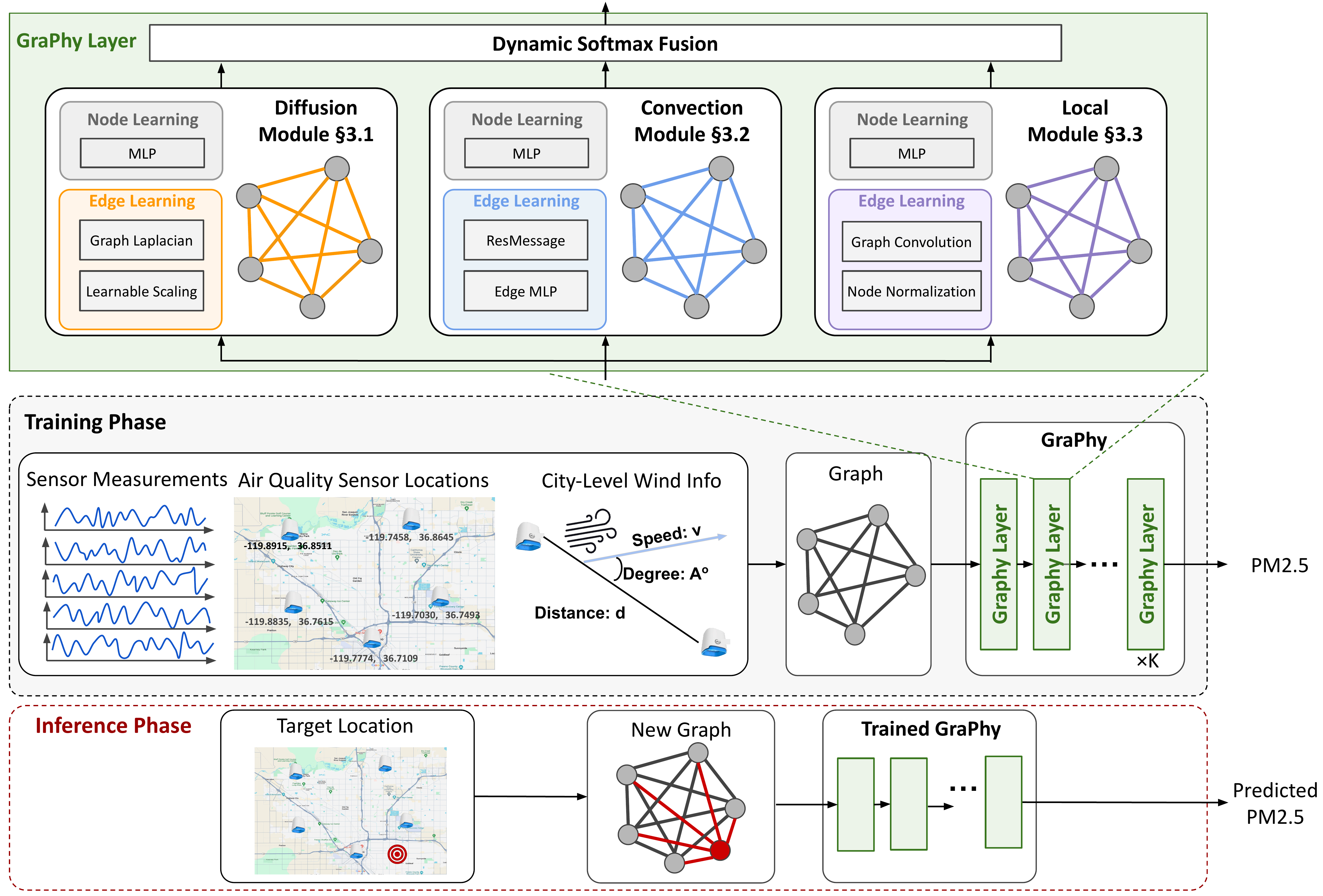}
    \vspace{-2ex}
    \caption{\name Overview. The model inputs sparse air quality sensor measurements, sensor location, and city-level wind information. These measurements determine the graph's node and edge features. 
    A GNN model consisting of multiple \name layers is used to model the air quality in the region. These physics-guided layers are customized to work with extremely low-resolution urban measurements. 
    We adopt complete graphs in each layer since graph structure learning is beyond the scope of this paper.
    The model is trained by recovering randomly masked measurements of existing sensors.
    To infer the measurement at an unmonitored location, a new graph is formed to include this location with dummy measurement input, and the model outputs the predicted measurements.
    }
    \label{fig:graphy overview}
\end{figure*}

\section{\name Design for Constrained Data} \label{sec:methodology}
In this work, we focus on the problem of air quality spatial interpolation, which aims to predict the air quality at any unmonitored location, given limited air quality sensors' measurements.
We introduce \name, a graph-based physics-guided machine learning model, to achieve fine-grained air quality estimation with constrained data of limited density and spatial resolution. 
As shown in Figure \ref{fig:graphy overview}, \name takes multimodal data as input, including air quality sensor measurements, sensors' geolocations, and city-level wind information. 
The graph is formed based on the sensors' geolocations.
Like traditional deep neural networks, \name is constructed by stacking layers, which we refer to as \name \textit{layers}.

In the \textbf{training phase}, for $N$ sensors in the target region, one will be randomly selected and masked with dummy variables.
The rest of the $N-1$ sensors' measurements will be modeled to recover the selected sensor measurements.
This process will be repeated $I$ times to generate a training dataset containing $I$ training samples.
For the $i$th repetition, we consider $x_i$ as the input data, which is the multimodal data including $N-1$ sensors' measurement, and $a_i$ as the ground truth data, which is the sensor measurements of the masked sensor.
Let $f_{\theta}$ denote the \name model with trainable parameters $\theta$.
The model is trained by minimizing the mean square error (MSE) loss: $\frac{1}{I}\sum_i^I (f_{\theta}(x_i) - a_i)^2$, where $I$ is the number of training samples.

In the \textbf{inference phase}, \name takes the unmonitored location with dummy sensor measurements (all zeros vectors) and the measurements from all the sensors (with no masking and removal) in the training phase as inputs. Please note that locations in the inference phase should be different from the locations used in the training phase.
Then, \name will form a new graph, connecting the unmonitored location and all locations with air quality data (i.e., locations in the training phase), and predict the air quality at this new location.

This training and inference design is essential, particularly when working with the \emph{existing dataset} for model development and evaluation.
First, because the dataset does not include sensor readings from actually unmonitored locations without ground-truth PM2.5 values, we train and evaluate the model by masking and recovering \emph{available} sensor data, which provide ground-truth for PM2.5 values. It is also crucial to divide the data into non-overlapping training and testing sets based on sensor locations to \emph{prevent data leakage and avoid overestimating model performance}.
Second, we randomly mask only one sensor at a time during training due to the limited number of sensors in regions with sparse air quality monitoring (e.g., SJV). This strategy \emph{maximizes contextual information for imputation} and \emph{minimizes distributional shifts between training and evaluation}, where $N - 1$ sensors are used as context during training, while all $N$ training sensors are used during evaluation.

To enhance the air quality modeling accuracy, we designed the \name layer to model the diffusion-convection processes that impact the air quality distribution, which is described by the diffusion-convection equation in Equation \ref{eq:CD-Eq}.
To better model the different physical processes described in this equation, we use three separate GNN modules, namely the Diffusion, Convection, and Local Modules, as shown in Figure \ref{fig:graphy overview}.

Each module has customized node and edge learning processes and a distinct node-edge interaction mechanism designed based on the corresponding terms in the equation.
They learn different features from the input data, and a Dynamic Softmax Fusion module, in which the fusing weights are dynamically determined by the input data distribution, will fuse the learned features to produce the layer output.
Each layer's output will become the following layer's input, and the last layer's output will be the final air quality prediction.
The forward process of the $k$-th \name layer is expressed in the equation below.
\begin{equation*}
    x_D^{(k)} = \text{GNN}_D^{(k)}(x^{(k-1)})
\end{equation*}
\begin{equation*}
    x_C^{(k)} = \text{GNN}_C^{(k)}(x^{(k-1)}) 
\end{equation*}
\begin{equation*}
    x_L^{(k)} = \text{GNN}_L^{(k)}(x^{(k-1)})
\end{equation*}
\begin{equation*}
    (w_D, w_C, w_L) = \text{Fusion}^{(k)}(x_D^{(k)}, x_C^{(k)}, x_L^{(k)})
\end{equation*}
\begin{equation*}
    x^{(k)} = w_D \cdot x_D^{(k)} + w_C \cdot x_C^{(k)} + w_L \cdot x_L^{(k)}
\end{equation*}

$x^{(k-1)}$ is the output features of the $(k-1)$-th \name layer and the input of  the $k$-th layer.
$\text{GNN}_D^{(k)}$, $\text{GNN}_C^{(k)}$, and $\text{GNN}_L^{(k)}$ are three GNN modules in the k-th layer, and $x_D^{(k)}$, $x_C^{(k)}$, and $x_L^{(k)}$ are the output features of each module, respectively.
Unlike the deterministic equations, the input data may impact the certainty of the output of multiple GNNs at different times.
Therefore, we use a dynamic softmax fusion module to combine the three output features.
This fusion module concatenates the output features $x_D^{(k)}$, $x_C^{(k)}$, and $x_L^{(k)}$ and inputs it to a MLP to generate three weights, 
$w_D$, $w_C$, and $w_L$.
The output feature of the layer, $x^{(k)}$, is the weighted sum of the output feature of each module.
The corresponding sensor data in the investigating time window $W$ is input into the model at the first layer as $x^{(1)}$.

We introduce the details of each GNN module in the sections below and explain how they can work with constrained data with limited spatial resolution.

\begin{figure*}[!t]
\minipage{0.33\textwidth}
  \includegraphics[width=\linewidth]{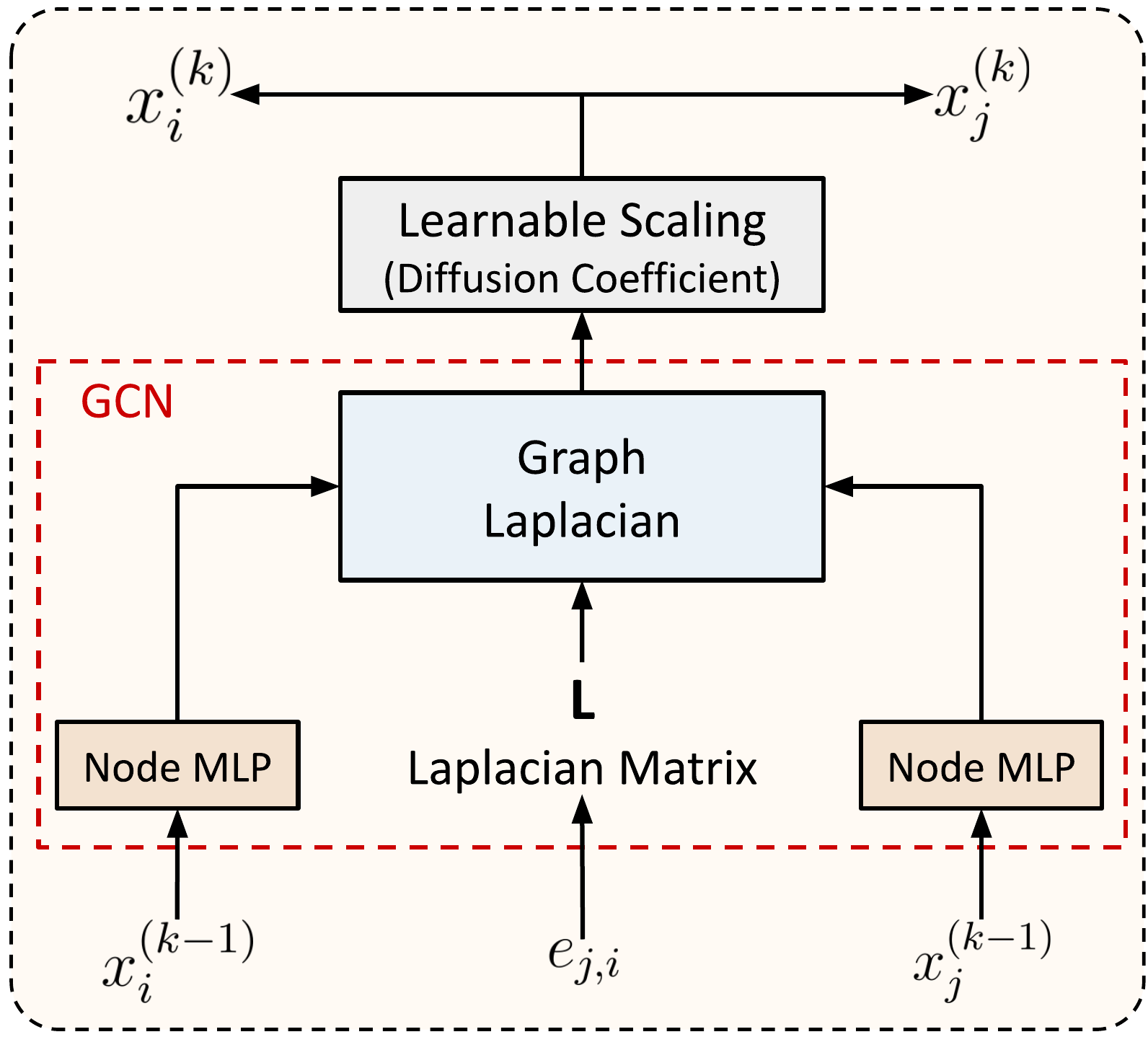}
  \vspace{-3ex}
  \caption{Diffusion Module} \label{fig:diffusion module}
\endminipage
\minipage{0.33\textwidth}
  \includegraphics[width=\linewidth]{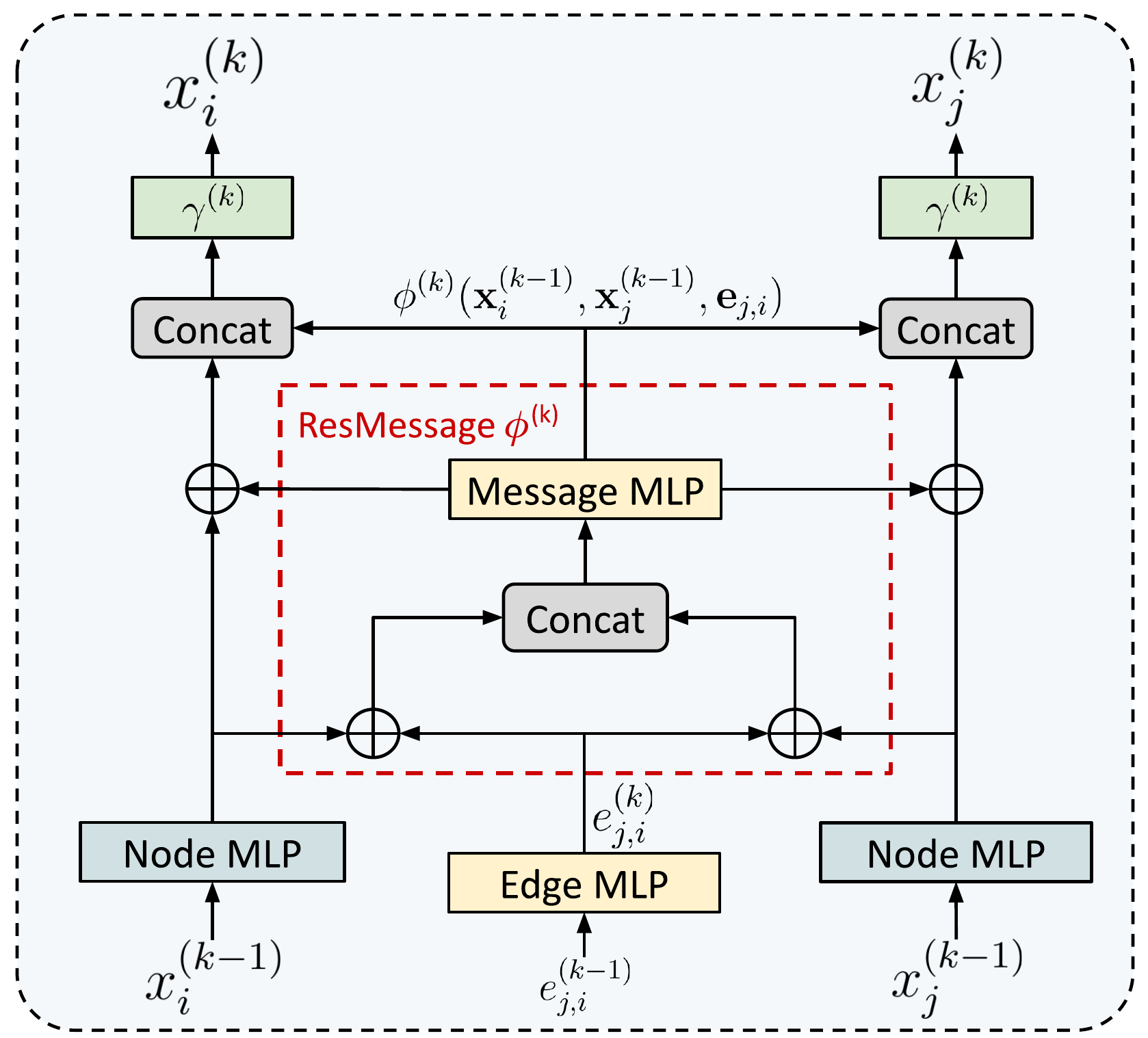}
  \vspace{-3ex}
  \caption{Convection Module}
  \label{fig:convection module}
\endminipage
\minipage{0.33\textwidth}
\includegraphics[width=\linewidth]{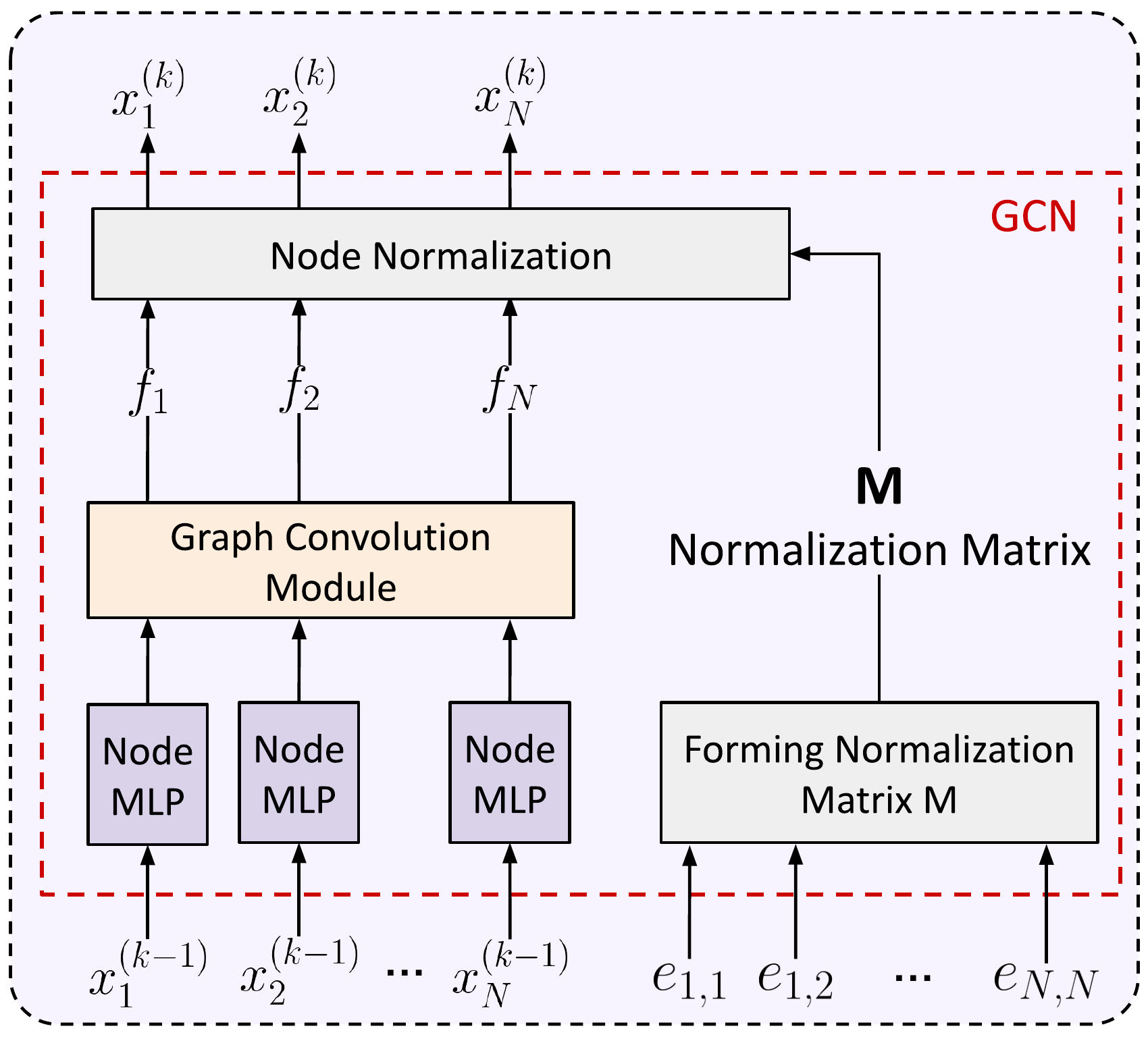}
  \vspace{-3ex}
\caption{Local Module}
\label{fig:local module}
\endminipage
\end{figure*}

\subsection{Diffusion Module} \label{subsec:diffusion}

The diffusion process describes the physical phenomenon of particles migrating from one region (of higher concentration) to another (of lower concentration), which is one of the most important aspects of describing air pollutants' movement. 
The state-of-the-art physics-guided deep learning model AirPhyNet adopts the graph diffusion, implemented as GCN, to mimic this diffusion process of air pollutants \cite{hettige2024airphynet}.
AirPhyNet has sensor measurements and sensor locations as input to the GCN, and the output is a feature vector.
They apply a fixed scaling value to this output feature vector to mimic the diffusion coefficient.
However, given the GCN output vector has unexplainable features, instead of using a fixed scalar to represent the diffusion coefficient, we use a learnable vector $l$ with the same dimension as the output vector to learn the separate effects of the diffusion coefficient on each feature.

Figure \ref{fig:diffusion module} shows the $k$-th layer of our diffusion module GNN applied on a pair of nodes $i$ and $j$. 
The nodes' input features ($x_{i}^{(k-1)}$ and $x_{j}^{(k-1)}$) will first undergo a multi-layer perception (MLP) to learn node-level representations.
We consider the physical distance $dist_{j,i}$ between sensors represented by these nodes as the edge feature  $e_{j,i} = \frac{1}{ dist_{j,i}}$.
Then, we construct a weighted adjacency matrix $A$ using these edge features, where $A(i,j) = e_{i,j}$.
The adjacency matrix $A$ is then used to define the graph Laplacian matrix further 
$L = I - D^{-\frac{1}{2}}AD^{-\frac{1}{2}}$,
where $D$ is the diagonal degree matrix.
This Laplacian matrix is then scaled to ensure computation stability as $L_D = 2 L/ \lambda_{\text{max}} - I$, where $\lambda_{\text{max}}$ is the maximum eigenvalue of the Laplacian matrix $L$. 
Then, we can mimic the physical diffusion process on a graph using the GCN introduced in Section \ref{subsec:diffusion on graph}:
$$\text{GNN}_D(X) = l \odot \sigma(L_DXW)$$
where $l$ is the learnable vector that represents the effects of the diffusion coefficient on output features and $\odot$ is element-wise multiplication. 

\subsection{Convection Module}
\label{subsec:convection}
Convection from winds is one of the major contributors to air pollutant dissemination \cite{stockie2011mathematics}. 
As a result, wind information, such as speeds and directions, has proven important in air quality modeling. 
AirPhyNet \cite{hettige2024airphynet} models such information based on the wind measurement at each air quality monitoring location.
However, for monitoring-constrained regions, such as the SJV, wind information is only available at the \textbf{city-level}, making the wind model in AirPhyNet infeasible for coarse-grained wind information.

We argue that modeling the \textbf{wind's impact on the air pollutant between two sensors} can achieve comparable wind modeling with only coarse-grained wind information. 
Therefore, we design the edge feature and GNN message-passing, focusing on the wind's effect on the air pollutants between two sensors.
As illustrated in Figure \ref{fig:graphy overview}, we construct the wind-based features $e_{j,i}$ using wind speed $w_{v}$, wind-sensor relative direction $w_{A}$, and distances $dist_{i,j}$ between two sensors as edge features for the convection module.
First, $w_{v}$ is important because it describes the magnitude and time delay of the wind effects on the pollutant.
Next, since we have a coarse resolution of $w_{v}$ information (city-level), $w_{A}$ is calculated as the directional similarity between the wind direction and the pair of sensors' geographical direction to model the wind effects between each pair of sensors.
To do so, we take the cosine of the angle between these two directions.
Finally, $dist_{i,j}$ is also essential because it determines the time delay of the wind effects between sensor $i$ and $j$.

Figure \ref{fig:convection module} shows the diagram of the convection module.
The node and edge MLPs first transform node features $x_{i}^{(k-1)}$ and $x_{j}^{(k-1)}$, and edge features $e_{j,i}^{(k-1)}$ to a comparable dimension.
The output of edge MLP are then used as the next layer's edge feature $e_{j,i}^{(k)}$.
We define the ResMessage function ($\phi^{(k)}$ in Equation \ref{eq:messagepassing}) that takes these node and edge features as input and output message features $\phi^{(k)}(x_i^{(k-1)}, x_j^{(k-1)}, e_{j,i})$.
The key trainable module of the ResMessage function is the Message MLP.
We form the input of the Message MLP as the concatenated features from each node, where the edge feature is added to the node feature to capture the edge's impact on the node independently. 
The output of the Message MLP is a message feature 
$\phi^{(k)}(x_i^{(k-1)}, x_j^{(k-1)}, e_{j,i})$, which represents the message sent from node $j$ to node $i$.
For node $i$, we concatenate the message feature with the sum of the node and message features to capture both the node-node interaction and the node-edge interaction.
This concatenated feature, $c_{i}$, is then input to the shared Update function $\gamma^{(k)}$, which is an MLP and outputs the next layer's node features $x_{i}^{(k)}$.
This shared Update function are applied to each node's concatenated features. 

\subsection{Local Module} \label{subsec:local}
The internal generation and removal of air pollutants within a region is the third key contributor to pollutant distribution, as described by the third term $R(s,t)$ in Equation \ref{eq:CD-Eq}.
This is extremely important for the SJV area, where the region faces multiple types of severe pollutant sources, such as agricultural activities, industrial emissions, and wildfire impacts, over different times.
Therefore, it is essential to identify and model these \textbf{instant local pollutant sources} and their impact on the air pollutant distribution. 

To capture sudden significant changes in the air pollution level, we use GNN-based interpolation to model the local source or sink of the air pollutants.
Here, we leverage the GCN to estimate the rough air pollutant level in the target location.
As demonstrated in Figure \ref{fig:local module}, for the $k$-th layer, each node's features $x_{i}^{(k-1)}$ will first be transformed into a latent space by a node MLP to get a node-level representation.
Then, this node-level representation is propagated to its neighboring nodes through graph convolution. 
The graph convolution module inputs all nodes' features and outputs the intermediate features $f_{i}\in \mathbb{R}^{c}$.
Meanwhile, the edge features $e_{j,i}$ are used to form a normalization matrix $M$, which is a diagonal matrix with element $M_{i,i} = 1 + \sum_{j \in \mathcal{N}(i)} e_{j,i}$ representing the degree of node $i$ with self-loops.
Then, the intermediate features $f_{i}$ of node $i$ will be normalized using the normalization matrix $M \in \mathbb{R}^{N \times N}$.
This normalization is performed by matrix multiplication $X^{(k)} = MF$, where $F \in \mathbb{R}^{N \times c}$ is the feature matrix containing the intermediate feature $f_{i}$ from each of the $N$ nodes.
$X^{(k)}$ final output feature matrix, where $i$-th row $x_i{(k)}$ is the output feature vector for node $i$ at $k$-th layer.

\section{Evaluation}\label{sec:eval}

We evaluate \name with real-world data collected by citizen science sensor networks and calibrated by Central California Asthma Collaborative's SJVAir \cite{sjvair2024website}.

\subsection{Dataset Preparation}
SJVAir acquires raw PM2.5 concentration measurements ($\mu g / m^3$) every 2 minutes \cite{purpleair2024documentation} from the deployed air quality sensors in the city.
Then, they conduct a customized calibration by fitting a linear regression model to the nearest reference air quality regulatory station.
The calibrated data is available from the website \cite{sjvair2024website}.

We target the City of Fresno, the largest city in the San Joaquin Valley, which often violates ambient air quality standards for particulate matter \cite{pinedo2024fresno}.
As the region is historically socioeconomically disadvantaged, limited measurement resources are available.
We select October 2023 to January 2024 as the studied period because, during the winter season, there is more wood burning for residential heating \cite{GridInfo_Fresno} and vehicle emission due to less efficient fuel combustion in cold weather \cite{fuel_economy}, which causes more severe air pollution.
As a result, the data demonstrates higher spatial heterogeneity due to more local PM2.5 sources.
We collect the time series of air quality measurements from 51 air quality sensors in Fresno during the studied period.
This time series data is pre-processed by averaging hourly following the procedures from previous works \cite{hsieh2015inferring, patel2022accurate, zheng2013u, hettige2024airphynet}.
Despite the preprocessing done by the SJVAir to ensure the quality and accuracy of the reported sensor measurements, the citizen science sensor network is inevitably missing data due to possible network or power losses.
In this work, we use the 41 sensors' data, ensuring that any missing data does not exceed a continuous duration of 1 hour. 
In this way, we guarantee that an hourly average measurement is always available.
For each of these sensors, we acquire 2928 hours of PM2.5 measurements.

In addition to the PM2.5 data, we prepare the City of Fresno wind dataset from the same studied period from Visual Crossing \cite{visualcrossing}, which provides city-level historical weather data.
The wind data contains hourly wind speed and wind direction information, which matches the sampling frequency of the PM2.5 dataset.
A detailed dataset summary is shown in Table \ref{tab:dataset details}.
\begin{table}[h!]
    \centering
    \caption{Dataset preparation summary}
    \vspace{-1ex}
    \label{tab:dataset details}
    \begin{tabular}{|c|c|c|}
        \hline
        \textbf{Data Type} & \multicolumn{2}{|c|}{\textbf{Statistics}} \\
        \hline
        \multirow{3}{*}{PM2.5} & \#.Sensors & 41 \\
         \cline{2-3}
         & Hours & 2928 \\
         \cline{2-3}
         & Time Spans & 10/01/2023-01/30/2024 \\
        \hline
        \multirow{3}{*}{Wind} & \#.Stations & 1 \\
         \cline{2-3}
         & Hours & 2928 \\
         \cline{2-3}
         & Time Spans & 10/01/2023-01/30/2024 \\
        \hline
    \end{tabular}
\end{table}

\subsection{Experiment Settings}
The models are implemented and trained in Python and Pytorch on a server with 2 NVIDIA GeForce RTX 4090 GPUs, 24 GB, and an AMD Ryzen Threadripper PRO 5955WX CPU.
The 41 sensors within the dataset are randomly split into training, validation, and testing set by a ratio of $28:4:9$.
The model is trained with a batch size of 32 samples, and we select Adam as the optimizer with a learning rate of $0.0001$ and $(\beta_1, \beta_2) = (0.9, 0.999)$.
The implementation of \name model contains five stacked \name layers with the hidden dimension set to 512.
Nine test sensors, excluded from training, are used to evaluate model performance. Their measurements serve as ground truth, while air quality estimates at their locations are predicted during inference. As described in Section \ref{sec:methodology}, data from 28 training sensors (referred to as \emph{context sensors}) are used as model inputs to estimate PM2.5 values at test sensor locations.

\begin{figure}[b!]
    \centering
    \includegraphics[width=\linewidth]{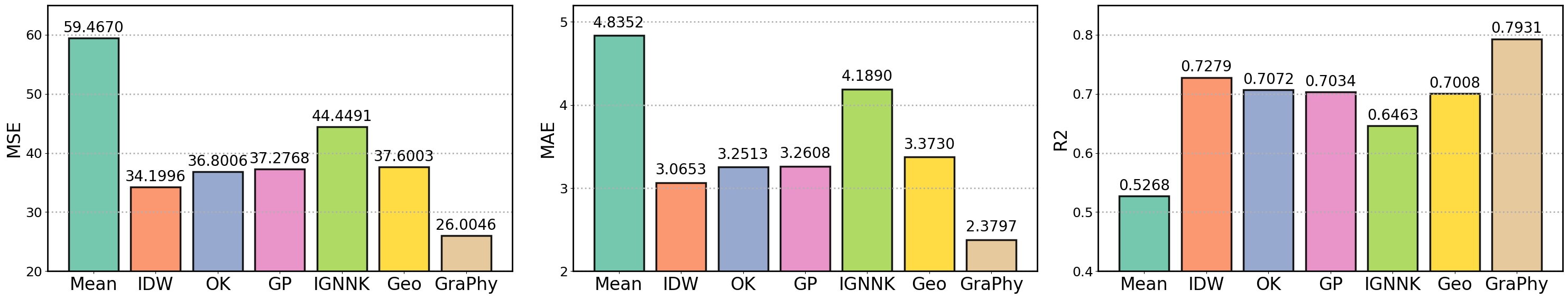}
    \caption{Prediction Accuracy of All Models on the Test Set.  \name achieves the \textbf{\emph{best}} performance, with the lowest MSE ($\downarrow$) and MAE ($\downarrow$) and the highest $R^2$ ($\uparrow$), while Mean Fill performs the worst. Compared to other models, \name reduces MSE by \textbf{24\%–56\%}, lowers MAE by \textbf{22\%–51\%}, and improves $R^2$ by \textbf{9\%–50\%}.}
    \label{fig:model_results_overall}
\end{figure}

\subsection{Baseline Models} \label{sec:baselines}
We consider several previous works as the baseline models, including three deterministic methods and two state-of-the-art deep learning-based methods.

\begin{enumerate}
\item \textit{Mean Fill} \cite{wei2024temporally}: Mean Fill is one of the simplest yet widely used imputation methods. In our case, we estimate the PM2.5 value at a target location and time step by averaging the PM2.5 values from all context sensors in the training set at the same time step. This approach treats all context sensors equally, assigning them the same weight. In our preliminary experiments, we also tested other basic imputation methods, such as Zero Fill and Median Fill, but their performance was inferior to Mean Fill. Therefore, we keep Mean Fill in the final results for comparison with other models.
\item \textit{Inverse Distance Weighting (IDW)} \cite{shepard1968two}: IDW is a distance-based non-trainable interpolation method used in geostatistics.
It calculates the air pollutant concentration at an unmonitored location as the weighted sum of the concentration at all monitored locations.
The weights are determined and normalized according to the inverse Euclidean distances between the target unmonitored and monitored locations.
\item \textit{Okriging} \cite{wackernagel2003ordinary}: Ordinary Kriging is another common geostatistics algorithm for spatial interpolation.
It assumes that data similarity decreases as the distance between data points increases and estimates a function to fit the distance-similarity relationship.
The interpolation is conducted by weighting the monitored air pollutant concentration according to the fitted function.
We implement Ordinary Kriging using the \textit{pykrige} package with a linear variogram model.
\item \textit{Gaussian Process (GP)} \cite{cheng2014aircloud}: Gaussian Process is a widely used method for spatial-temporal interpolation.
It is a non-parametric method that uses Bayesian Statistics to model the air quality distribution under a multivariate Gaussian assumption.
We implement the algorithm using the \textit{gpytorch} package in Python.
We use a constant mean function and a Matern Kernel with $\nu = 0.5$ for the covariance function.
\item \textit{IGNNK} \cite{wu2021inductive}: IGNNK is a GNN-based model for inductive kriging.
It uses a node-level multilayer perceptron (MLP) to project node features to a high-dimensional space and leverages a distance-based adjacency matrix and the Chebyshev polynomial to approximate the graph convolution operation.
We follow the original implementation and hyperparameter settings provided in the GitHub codebase \cite{ignnk2024}.
\item \textit{Geo-LSTM} \cite{ma2019temporal}: Geo-LSTM is a deep learning model with geospatial filters for air quality mapping.
It uses a geospatial filtering module to select a $K$ number of nearby sensors and a Long-Short-Term Memory (LSTM) neural network to learn the time series data.
Since the codebase is not published by the author, we implement the model according to the methodology in the original paper.
\end{enumerate}

Since AirPhyNet \cite{hettige2024airphynet} does not apply to city-level wind data, it is not included in the baseline comparison for fairness.
We implement it to compare and illustrate our design's robustness to constrained data.
Since the AirPhyNet model is designed for forecasting tasks, we implement a variant (AirPhyNet-V) where the ODE solver is removed, and new nodes can be added to existing graphs for interpolation tasks.
We conduct an interpolation experiment using AirPhyNet-V and get a negative $R^2$ of $-1.9$, which indicates that the model fails to learn information from the dataset.
The corresponding mean absolute error is 15.6571 $\mu g / m^3$, which is significantly higher than other baselines and \name, which indicates their strong dependency on fine-grained meteorology data.
This is mainly because the coarse-grained wind data makes wind features identical on all nodes, and disables the message passing in the convection module (Appendix A.2 in \cite{hettige2024airphynet}).
As a result, it is hard for unmonitored locations to gather information from monitored locations, which leads to a random guessing prediction and degraded accuracy.

\subsection{Evaluation Metrics}

We evaluate interpolation accuracy using mean squared error (MSE), mean absolute error (MAE), and the R-Square value ($ R^2$).
In our evaluation, we consider the testing set sensor $i$ to be the unmonitored location.
The model prediction is denoted as $\Tilde{x}_i$, and the ground truth measurement is denoted as $x_i$. Also, $\bar{x}$ denotes the average of all testing ground-truth measurements $x_i$'s (i.e., $\bar{x}=\sum_{i=1}^{M} x_i / M$), where $M$ is the total number of testing samples.
The evaluation metrics are calculated as follows:

\begin{align*}
    MSE &= \frac{1}{M} \sum_{i=1}^M (x_i - \Tilde{x}_i)^2 &
    MAE &= \frac{1}{M} \sum_{i=1}^M |x_i - \Tilde{x}_i| &
    R^2 &= 1 - \frac{\sum_{i=1}^{M} (x_i - \Tilde{x})^2}{\sum_{i=1}^{M} (x_i - \bar{x})^2}
\end{align*}

\subsection{Results Analysis} \label{subsec:model performance}
We analyze the 26352 (2928 hours $\times$ 9 locations) PM2.5 estimations for all the baseline models and \name.
For the \emph{neural network-based approaches} (IGNNK, Geo-LSTM, and \name), we repeat the experiments five times with different random seeds for model initialization and permutation of training samples. To compute the evaluation metrics, we first average the predictions from five models trained with different seeds (i.e., ensemble averaging). 
We then calculate the evaluation metrics based on this averaged prediction and the ground-truth values. 
This approach aligns with real-world deployment, where a single prediction is required.

\subsubsection{Overall Model Accuracy}
Figure \ref{fig:model_results_overall} compares MSE, MAE, and $R^2$ across \name and baseline models. We can see \name achieves the \emph{best} performance on all three metrics, while Mean Fill performs the worst. Additionally, \name significantly outperforms other baselines ( $p<0.0001$ with the paired t-test), with 24\%–56\% reduce on MSE, 22\%-51\% decrease on MAE, and 9\%-50\% improvement on $R^2$. Unlike Mean Fill, all other models incorporate the spatial similarity between context sensor locations and target locations, highlighting the importance of distance information. Interestingly, deep learning-based baselines (i.e., IGNNK and Geo-LSTM) perform worse than simpler statistical models (i.e., IDW, Okriging, and GP), which needs further investigation.

\subsubsection{Model Accuracy by PM2.5 Values}

\begin{figure}[htbp]
    \centering
    \begin{subfigure}[b]{0.40\textwidth}
        \centering
        \includegraphics[width=\linewidth]{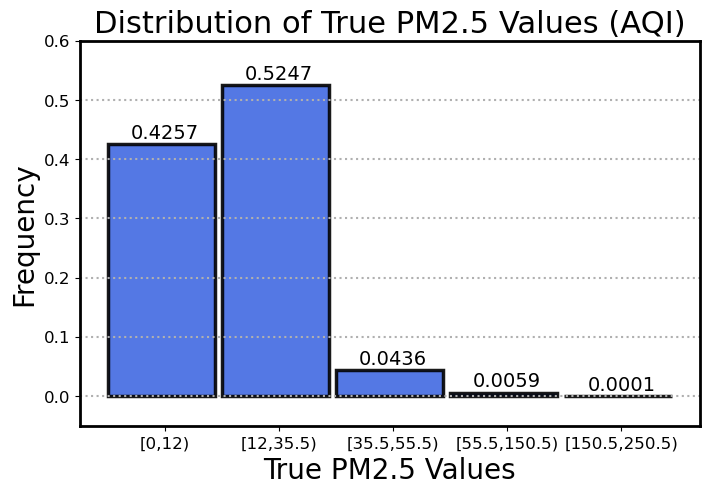} 
        \caption{}
        \label{fig:gt_dist_aqi}
    \end{subfigure}%
    \begin{subfigure}[b]{0.40\textwidth}
        \centering
        \includegraphics[width=\linewidth]{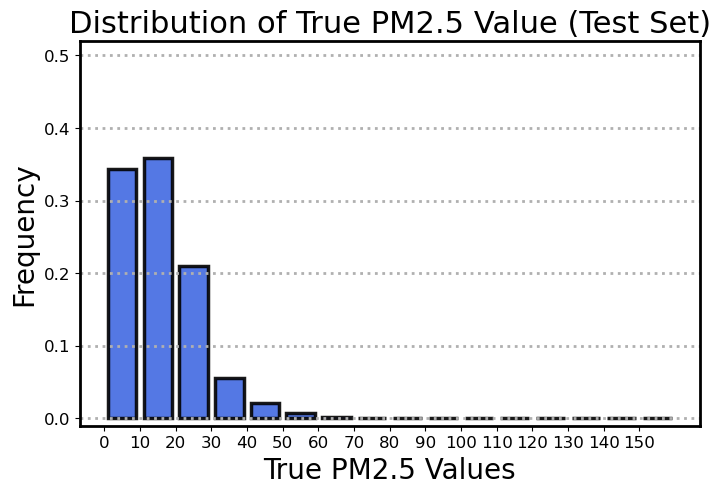} 
        \caption{}
        \label{fig:gt_dist_test}
    \end{subfigure}%
    \hfill
    \begin{subfigure}[b]{0.40\textwidth}
        \centering
        \includegraphics[width=\linewidth]{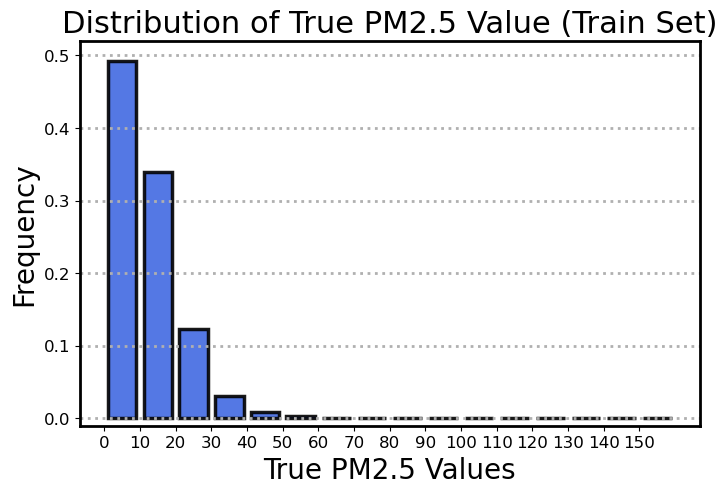} 
        \caption{}
        \label{fig:gt_dist_train}
    \end{subfigure}%
    \begin{subfigure}[b]{0.40\textwidth}
        \centering
        \includegraphics[width=\linewidth]{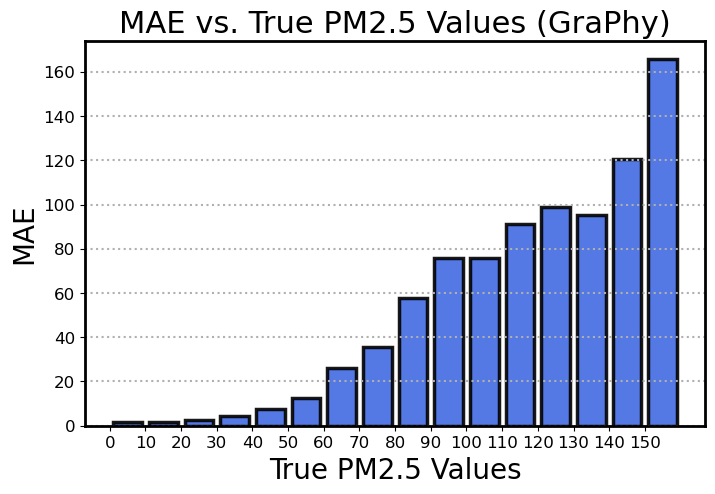} 
        \caption{}
        \label{fig:mae_gt_graphy}
    \end{subfigure}

    \caption{Distribution of Ground-Truth PM2.5 Values and MAE Performance of \name. (a) Ground-truth PM2.5 distribution based on the Air Quality Index (AQI). (b) Ground-truth PM2.5 distribution in the test set. (c) Ground-truth PM2.5 distribution in the training set. (d) MAE of \name across ground-truth PM2.5 values.}
    \label{fig:four_subfigures}
\end{figure}

We further analyze the computational results by examining the distribution of ground-truth PM2.5 values of the test set and evaluating model performance across different PM2.5 value bins. Performance is measured in terms of MAE, considering all sensors and time steps where the ground-truth PM2.5 values fall within each bin.  Figure \ref{fig:gt_dist_aqi} presents the distribution of ground-truth PM2.5 values based on the Air Quality Index (AQI) \footnote{\url{https://www.epa.gov/sites/default/files/2016-04/documents/2012_aqi_factsheet.pdf}}, indicating that most cases fall within the "good" (PM2.5: 0-12) and "moderate" (PM2.5: 12-35.5) air quality categories. To assess model performance at a finer granularity, we further divide the ground-truth values into smaller bins with a size of 10, as shown in Figure \ref{fig:gt_dist_test}. The distribution is unbalanced and skewed toward lower values.

\begin{figure}[!t]
    \includegraphics[width=\linewidth]{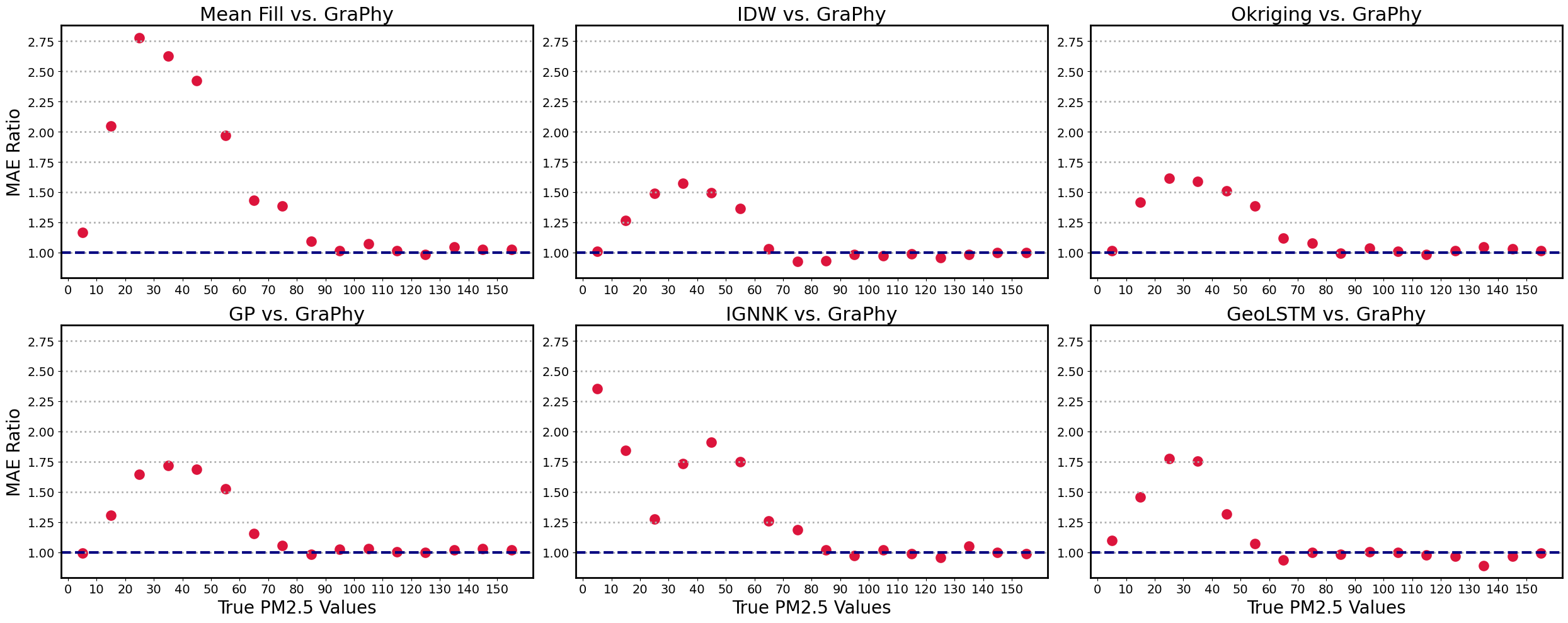}
    \vspace{-3ex}
    \caption{Ratio of the MAE from the baseline models to that of the \name model for each ground-truth PM2.5 bin. A ratio greater than 1.0 indicates that the baseline model has a higher error than \name for the corresponding bin.}
    \vspace{-2ex}
    \label{fig:mae_ratio_gt}
\end{figure}

Figure \ref{fig:mae_gt_graphy} presents the MAE for each ground-truth bin of the proposed \name model. The model exhibits higher errors for bins with higher ground-truth PM2.5 values, which is expected given that these values are much less frequent in the training set (as shown in Figure \ref{fig:gt_dist_train}). Figure \ref{fig:mae_ratio_gt} shows the ratio of the MAE from the baseline models (described in \ref{sec:baselines}) to that of the \name model for each bin. A ratio greater than 1.0 indicates that the baseline model has a higher error than \name for the corresponding PM2.5 bin. The results indicate that \name outperforms all baseline models for lower PM2.5 values, particularly in the 10–60 range. However, for higher values ($\geq 80$), the baseline models perform similarly to \name. When combined with the distribution shown in Figure \ref{fig:gt_dist_test}, these findings suggest that \name’s overall superior performance is primarily driven by its accuracy on lower PM2.5 values, which account for the majority of the test data.

\subsubsection{Model Accuracy by Spatial Heterogeneity} \label{sec:acc_sh}

Spatial heterogeneity quantifies the variation in PM2.5 levels across different sensor locations. Higher spatial heterogeneity suggests more complex physical conditions, such as local pollutant sources, intricate wind patterns, or other factors affecting pollutant distribution. These complexities make it more challenging for the model to predict a sensor's values based on neighboring sensors. Therefore, we further analyze model accuracy in terms of spatial heterogeneity.

Formally, spatial heterogeneity (SH) at a given time step $t$ is defined as the variance of air quality measurements across all sensors at that time step: 

\begin{equation}
\label{eq:sh}
    \text{SH}_t = \frac{1}{N-1} \sum_{j=1}^N (x_{j,t} - \bar{x}_{t})^2
\end{equation}
where $\bar{x}_{t}$ is the mean air quality measurement at time $t$, and $x_{j,t}$ is the measurement from sensor $j$. Note that this metric is computed using all 41 sensors in the dataset, not just those in the test set.

To evaluate model performance based on spatial heterogeneity, we first visualize the distribution of spatial heterogeneity across all time steps. We then categorize model performance, measured by MAE, into different spatial heterogeneity bins and compare baseline models with our \name within each bin.

\begin{figure}[htbp]
    \centering
    \begin{subfigure}[b]{0.33\textwidth}
        \centering
        \includegraphics[width=\linewidth]{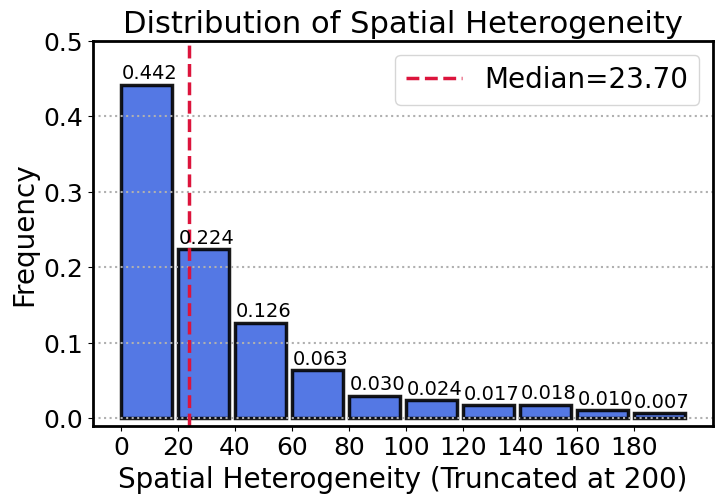} 
        \caption{}
        \label{fig:sh_dist}
    \end{subfigure}%
    \begin{subfigure}[b]{0.33\textwidth}
        \centering
        \includegraphics[width=\linewidth]{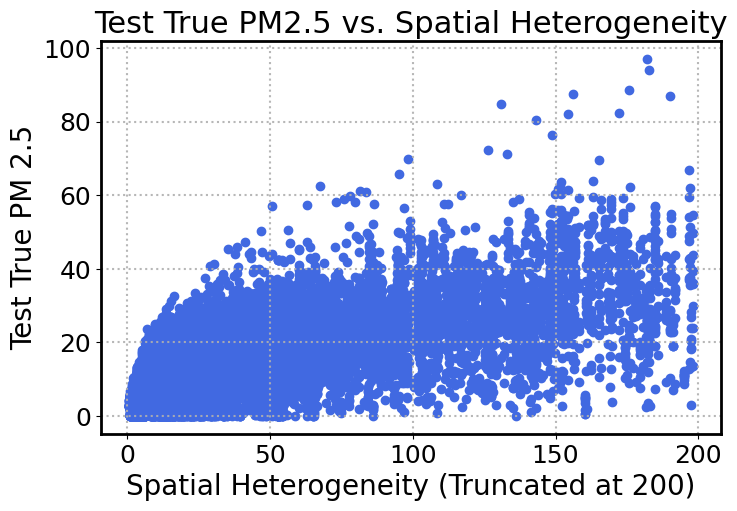} 
        \caption{}
        \label{fig:sh_gt_test}
    \end{subfigure}%
    \begin{subfigure}[b]{0.33\textwidth}
        \centering
        \includegraphics[width=\linewidth]{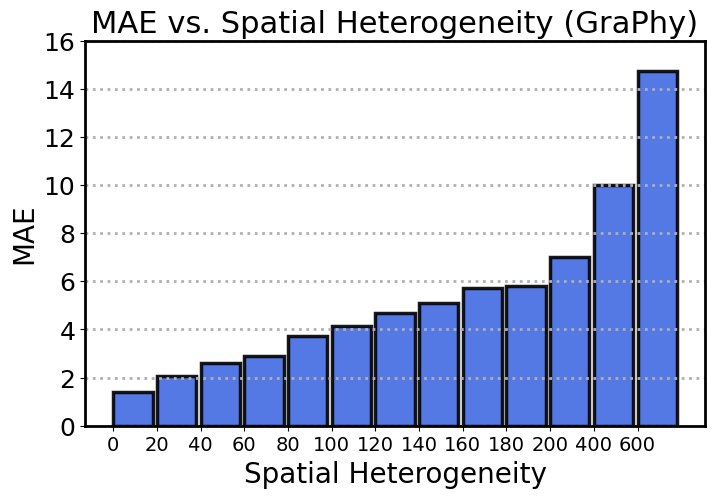} 
        \caption{}
        \label{fig:mae_sh_graphy}
    \end{subfigure}%

    \caption{Distribution of Spatial Heterogeneity values and MAE Performance of \name. (a) Spatial heterogeneity distribution across all time steps (results are truncated at 200 for better visualization). (b) Spatial heterogeneity values of each test sample and its corresponding ground-truth PM2.5 value (results are truncated at 200 for better visualization). (c) MAE of \name across spatial heterogeneity values. Last three bins have larger bin size than previous ones (200 vs. 20).}
    \label{fig:dist_spatial_heterogeneity}
\end{figure}

\begin{figure}[htpb!]
    \includegraphics[width=\linewidth]{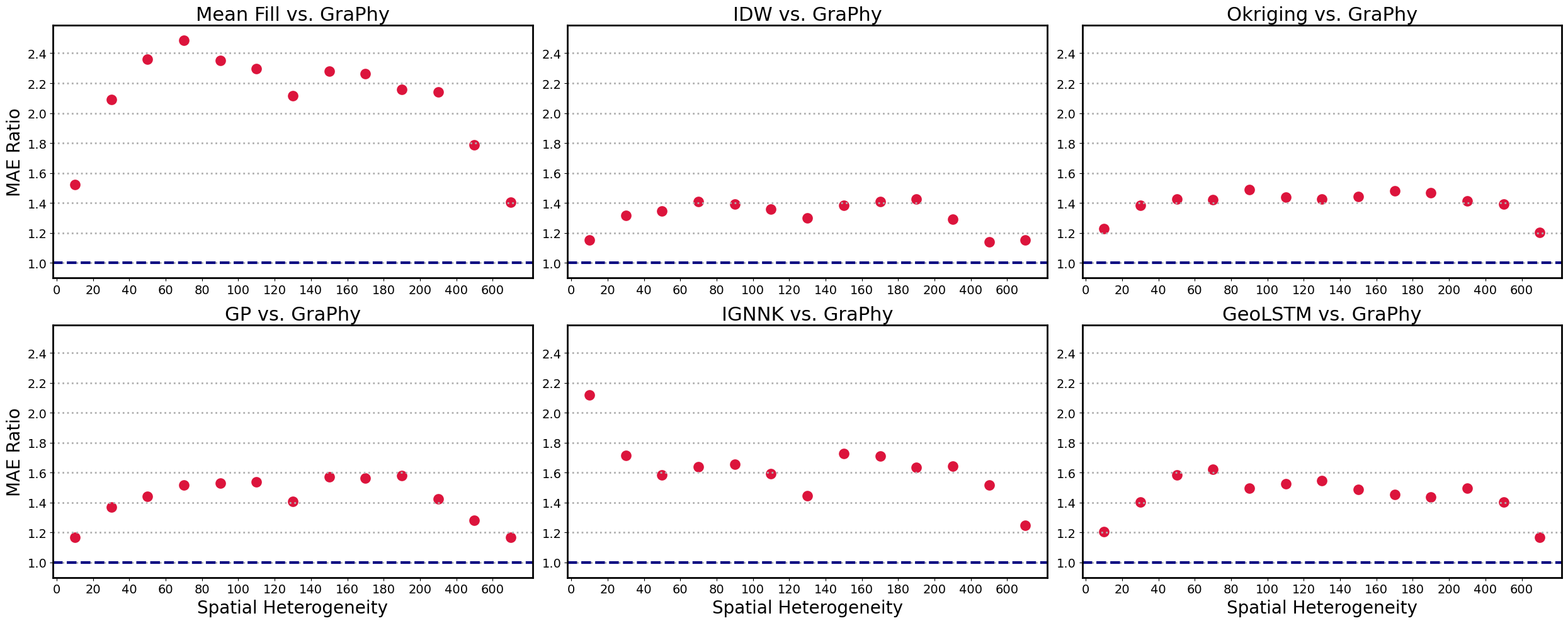}
    \vspace{-3ex}
    \caption{Ratio of the MAE from the baseline models to that of the \name model for each spatial heterogeneity bin. A ratio greater than 1.0 indicates that the baseline model has a higher error than \name for the corresponding bin.}
    \vspace{-2ex}
    \label{fig:mae_ratio_sh}
\end{figure}

Figure \ref{fig:sh_dist} shows that spatial heterogeneity in Fresno during the examined period (10/01/2023 to 01/30/2024) is generally low, with fewer instances of high heterogeneity. Figure \ref{fig:mae_sh_graphy} presents the MAE of \name across different spatial heterogeneity levels. Note that the last three bins have a different width than the others for better visualization without affecting the final conclusion. The results indicate that \name has higher errors when PM2.5 levels are more heterogeneous, making it more challenging to predict a sensor’s value based on others. Figure \ref{fig:mae_ratio_sh} displays the ratio of the MAE of baseline models to that of \name, where values above 1.0 indicate that the baseline model has higher errors in the corresponding spatial heterogeneity bin. The results show that all baseline models consistently exhibit higher errors across all spatial heterogeneity bins compared to the proposed model.

Moreover, while model performance follows a similar trend with respect to both ground-truth PM2.5 (Figure \ref{fig:mae_gt_graphy}) and spatial heterogeneity (Figure \ref{fig:mae_sh_graphy}), with higher errors at higher values, spatial heterogeneity does not necessarily correspond to higher ground-truth PM2.5 values. This is verified in Figure \ref{fig:sh_gt_test}, where each point represents a single testing sensor at a specific time step. The key distinction is that spatial heterogeneity reflects the \emph{differences} in PM2.5 values between sensor locations at a given time, whereas ground-truth PM2.5 is based on \emph{absolute} values. This demonstrates that higher prediction errors can be attributed separately to higher ground-truth PM2.5 and greater spatial heterogeneity, highlighting the necessity of analyzing both factors.

\subsubsection{Robustness against Sensor Density.}
When deploying the system in real-world scenarios, sensor density (i.e., the number of sensors per square mile) varies. In more rural areas, sensor density may be lower than the current level in the City of Fresno, making accurate PM2.5 imputation more critical. Therefore, we evaluate how robust our model is to lower sensor density and compare it with all the baseline models to understand our model's robustness and limitations better.

To simulate varying sensor densities, we randomly remove $20\%$, $40\%$, $60\%$, and $80\%$ of context sensors (i.e., 28 sensors in the training set), resulting in densities of $0.22/\text{mi}^2$, $0.16/\text{mi}^2$, $0.11/\text{mi}^2$, and $0.06/\text{mi}^2$.\footnote{After removal, 23, 17, 12, and 6 sensors remain, respectively, in Fresno’s 104.8 mi² area.} For each sensor density, we repeat the random removal process five times, generating five different splits with distinct subsets of remaining sensors. This ensures diverse sensor location distributions and prevents biases from specific sensor removal patterns, improving the generalizability of our results. \footnote{Exhaustively testing all possible sensor removal combinations is computationally infeasible, especially when the number of removed sensors is small. For example, removing six sensors (leaving 22) from a set of 28 results in $\text{C}_{28}^{6}$ = 376,740 possible combinations.} We compute the MAE for each split and report the average MAE across all five splits for each model and sensor density.

\begin{table}[]
\caption{Average MAE across five random splits for each sensor density. The density of 0.27/mi² corresponds to retaining all original context sensors. \textcolor{red}{Red} highlights the best performance for each density, while \textcolor{blue}{Blue} indicates graph-based models with the worst performance at the lowest density.}
\vspace{-2ex}
\label{tab:sensor_density}
\resizebox{0.5\textwidth}{!}{%
\begin{tabular}{lccccc}
\hline
               & \multicolumn{5}{c}{\textbf{Sensor Density} {\small($\text{\# of sensors}/\text{mi}^2$)}}                                                                                    \\ \cline{2-6} 
\textbf{Model} & \textbf{0.27} & \textbf{0.22} & \textbf{0.16} & \textbf{0.11}                          & \textbf{0.06}                          \\ \hline
Mean Fill \cite{wei2024temporally}      & 4.8352       & 4.7045        & 4.7450        & 4.5837                                 & 5.3721                                 \\
IDW  \cite{shepard1968two}           & 3.0653       & 3.1137        & 3.3569        & {\color[HTML]{FF0000} \textbf{3.3214}} & 5.3247                                 \\
Okriging \cite{wackernagel2003ordinary}      & 3.2513       & 3.4137        & 3.4719        & 3.5623                                 & 4.5151                                 \\
GP \cite{cheng2014aircloud}            & 3.2608       & 3.2729        & 3.4726        & 3.4957                                 & {\color[HTML]{FF0000} \textbf{4.4955}} \\
IGNNK \cite{wu2021inductive}    & 4.1890       & 4.2935        & 4.3029        & 4.4345                                 & {\color[HTML]{3531FF} 8.0749}          \\
GeoLSTM \cite{ma2019temporal}   & 3.3730       & 3.2508        & 3.4953        & 3.6976                                 & 5.1649                                 \\
\name \emph{(Ours)} & {\color[HTML]{FF0000} \textbf{2.3797}} & {\color[HTML]{FF0000} \textbf{2.5133}} & {\color[HTML]{FF0000} \textbf{2.9457}} & 4.3160 & {\color[HTML]{3531FF} 8.0465} \\ \hline
\end{tabular}%
}
\end{table}

Table \ref{tab:sensor_density} presents the average MAE results across different sensor densities, with red indicating the best (lowest MAE) performance for each density. The density of $0.27/\text{mi}^2$ corresponds to all context sensors (28 sensors) with no removal. The table shows that when sensor density is at least $0.16/\text{mi}^2$, \name outperforms all baseline models. However, at lower densities ($0.11/\text{mi}^2$ and $0.06/\text{mi}^2$), IDW and GP achieve the best performance, respectively. The decline in \name's performance at lower densities is likely due to a distribution shift between the training set, which has higher sensor density, and the test set we used here, where many context sensors have been removed. Interestingly, Mean Fill performs better at lower densities ($0.22/\text{mi}^2$ and $0.16/\text{mi}^2$) than at the highest density ($0.27/\text{mi}^2$), likely due to oversmoothing where mean imputation fails to capture data variations when too many context sensors are present. Additionally, we observe that graph-based models (IGNNK and \name) struggle with extremely low sensor density ($0.06/\text{mi}^2$), as indicated in blue. This occurs probably because the sparse graphs caused by extremely low sensor density contain too many noisy or irrelevant edges, which GNNs may amplify their impacts during global aggregation \cite{ye2021sparse}.

\subsubsection{Effect of Model Size}
We further analyze the trade-off between model complexity (size) and accuracy (measured by MAE and \( R^2 \)) by training models on the same dataset with varying hyperparameter settings. The number of stacked \name layers and the hidden dimension of each layer directly influence model complexity, leading to different model sizes.

We investigate three model configurations: 
\begin{enumerate}
\item \textit{Small} (\name-S): three stacked layers with a hidden dimension of 128; model size of 3.3MB.
\item \textit{Medium} (\name-M): four stacked layers with a hidden dimension of 256; model size of 20MB.
\item \textit{Large} (\name-L): five stacked layers with a hidden dimension of 512; model size of 105MB.
\end{enumerate}

\begin{figure}[!t]
    \centering
    \begin{subfigure}[b]{0.45\textwidth}
        \centering
        \includegraphics[width=\linewidth]{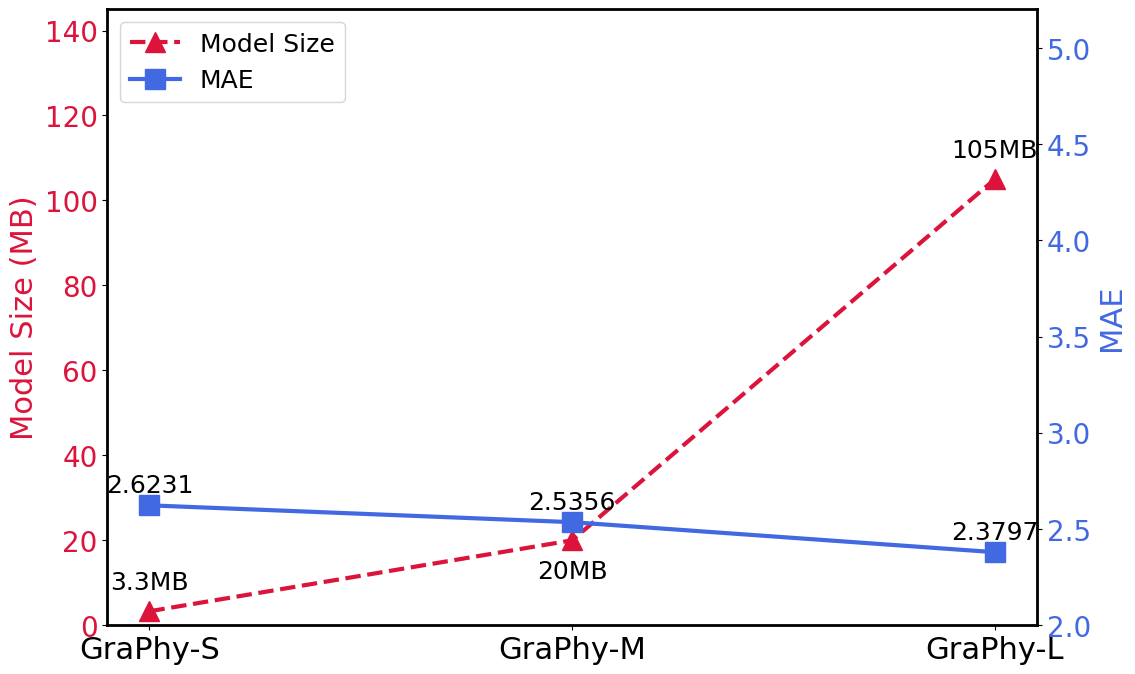} 
        \caption{Mean Absolute Errors across different model sizes.}
        \label{fig:model_size_mae}
    \end{subfigure}%
    \begin{subfigure}[b]{0.45\textwidth}
        \centering
        \includegraphics[width=\linewidth]{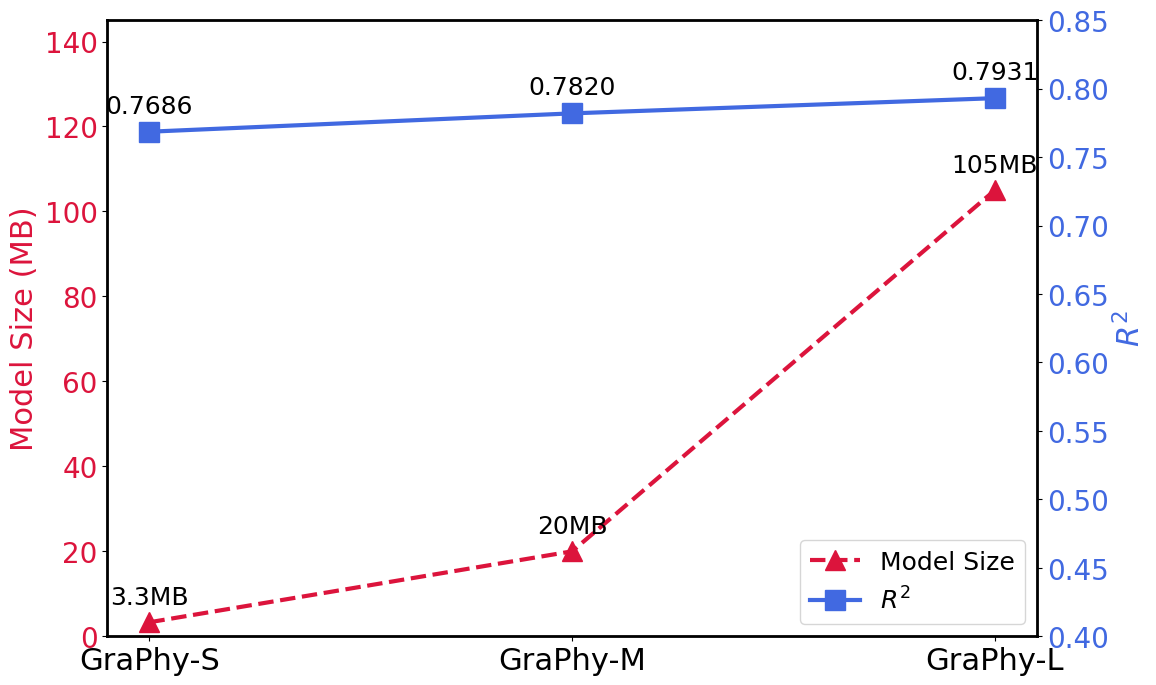} 
        \caption{$R^2$ across different model sizes.}
        \label{fig:model_size_r2}
    \end{subfigure}%

    \caption{Performance Across Different Model Sizes.  The left y-axis represents model size (\textcolor{red}{Red} line), while the right y-axis shows MAE or $R^2$ of the test results (\textcolor{blue}{Blue} line). GraPhy-S achieves an $R^2$ only 1.74\% lower than GraPhy-L, while its MAE is 10\% lower, despite being 33 times smaller. This demonstrates our model's adaptability to devices with varying computational power while maintaining comparable performance.}

    \label{fig:model_size}
\end{figure}

Figure \ref{fig:model_size} compares model size and performance across three configurations. The red line represents model size, while the blue line shows testing MAE and $R^2$ values. In general, larger and more complex models achieve higher accuracy. However, in our case, \name-S maintains strong performance, with only a 2\% decrease in $R^2$ and a 10\% decrease in MAE compared to \name-L, while reducing model size by 33×. This demonstrates the model’s adaptability for deployment on devices with varying computational capabilities (e.g., mobile devices) while maintaining comparable accuracy.

\section{Related Work}\label{sec:related}
This section introduces related work, including prior work on urban air quality sensing, large-scale simulation-based models, data-driven models, and physics-guided hybrid models.

\subsection{Urban Air Quality Monitoring}
Urban air quality monitoring focuses on measuring and analyzing the concentration of air pollutants in urban environments.
Common approaches to acquire the monitoring measurements, including Beta-Attenuation Method and Gravimetric Method used by regulatory stations \cite{gobeli2008met, yanosky2001comparison}, satellite imaging for remote sensing \cite{martin2008satellite, engel2004recommendations}, and distributed sensor network \cite{cheng2014aircloud, hsieh2015inferring, chen2020adaptive, liu2024mobiair}.
We focus on modeling air quality data acquired from the sensor network due to (1) their high sampling rate compared to the regulatory station, which allows high temporal resolution modeling and promptly detects instant pollutants, and (2) their real-time ground-based measurement compared to satellite remote sensing, which is often impacted by the cloud and delayed due to transmission latency.

Sensor network-based monitoring often falls into two categories: stationary citizen science sensor network \cite{sjvair2024website, purpleair} and vehicle-based mobile sensor network \cite{liu2024mobiair, devarakonda2013real}.
Sensors on vehicle-based platforms are often expensive, with up to USD 7,000 quote prices.
Furthermore, they often require dedicated vehicles to cover an extensive area of the city, which is not an efficient option for suburban areas with limited public transportation or ride-sharing. We focus on the stationary citizen science sensor network data and aim to interpolate the air quality to unmonitored locations, given the known pollutants' concentration at monitored locations. 
Our design targets enhancing spatiotemporal modeling with sparse measurements from the monitoring-constrained regions.

\subsection{Air Quality Spatiotemporal Modeling}
The spatiotemporal modeling of air quality monitoring data has been widely explore. Here, we summarize two major categories of solutions -- physics-based and data-driven models.

\paragraph{Physics Simulation Models}
Based on physics processes, simulation models usually input air pollutant source data and area meteorology data and output the area concentration maps.
For example, SLAB uses a steady-state plume model to simulate the dispersion of air pollutants with time-dependent release \cite{ermak1990user}.
CAMx simulates the dispersion of PM pollutants based on the real-world physicochemical process \cite{favez2010inter}.
CALPUFF simulates the dispersion of pollutants over long distances and complex terrains \cite{scire2000user}.
Although they require only a small amount of measurement data as input, they often do not model dynamic factors in the physical world, leading to errors in areas far from the measurements \cite{pielke1998use}. 

\paragraph{Data-Driven Models}
With the development of urban-scale air quality sensor networks and the recent advances in machine learning algorithms, many data-driven models have been explored to interpolate the area air quality.
For example, \citeauthor{patel2022accurate} uses a non-stationary Gaussian Process with specialized categorical and periodic kernels to conduct interpolation \cite{patel2022accurate}.
Geo-LSTM is a deep learning model that models air pollutants of an area using nearby sensor measurements with Long-Short-Term-Memory(LSTM) for temporal modeling \cite{ma2019temporal}.
IGNNK takes the sensor network as a graph with nodes and edges and treats unmonitored locations as virtual nodes on a graph for modeling \cite{wu2021inductive}. 
The model treats air quality sensor measurements of node features, and an inductive graph neural network is used to predict the air quality on virtual nodes through graph convolution.
However, these models often require a large amount of training data to model the physical world effectively, and our \name outperforms these models in our evaluation analysis. 

\subsection{Physics-Guided Machine Learning (PGML)} 


Recent research integrates physics-based laws and knowledge into machine learning algorithms to guide learning and constrain outputs, reducing training data requirements and mitigating overfitting. This approach is particularly suitable for physical data from IoT systems, where data availability is often limited.

\paragraph{PGML for Air Quality Estimation.} 
For air quality estimation, the convection-diffusion differential equation \cite{stocker2011introduction} is commonly used to model particle diffusion in the atmosphere. For example, HMSS \cite{chen2020adaptive} predicts spatial air quality using Gaussian Processes and models discrete-time changes by solving the convection-diffusion equation. This method combines the strengths of data-driven models, which capture influential factors not represented in physics-based models, with the physical constraints of the convection-diffusion equation, resulting in predictions three times more accurate than baseline models. Similarly, AirPhyNet \cite{hettige2024airphynet} integrates the convection-diffusion equation with Graph Neural Networks (GNNs), leveraging the Graph Laplacian operation to mimic physical convection and diffusion processes for improved air quality forecasting. Our \name is inspired by AirPhyNet, but is specifically designed for interpolation tasks with sparse measurements and high spatial heterogeneity.

\paragraph{PGML for IoT Sensor Data.} 
Based on the sensing principle and the data characteristics impacted by different physics laws, many PGML approaches have been explored.
For example, Zhang et al. \cite{zhang2020physics} apply dynamic laws to constrain CNN outputs when modeling buildings' structural seismic responses using motion sensor data. 
Daw et al. \cite{daw2022physics} incorporate fluid dynamics equations into neural network loss functions, effectively predicting lake water temperature. 
Rohal et al. \cite{rohal2024don} design a circuit-guided CNN to model the multiplex pressure sensing circuit's cross-talk effectively.
Mirshekari et al. \cite{mirshekari2020step} introduce a structure-aware model transfer approach that projects human-induced vibration data in a feature space where the structural effects are minimized.
Yu et al. \cite{yu2021vibration} leverage the environmental measurements to determine the optimal sensor placement via Thompson Sampling.
Hu et al. \cite{hu2022vma} propose a model transfer framework that utilizes the physical properties of sensing modalities to mitigate multi-factor domain shifts.
In addition to data distribution shift, physics measurements are integrated with causal intervention techniques to mitigate physical confounders' impact \cite{hu2022ciphy}.
Our model analysis in Section \ref{subsec:model performance} indicates our current model may be impacted by potential data distribution shifts and bias caused by season or city differences, and these PGML-based methods can be used with our GNN model to improve its robustness further.

\subsection{Learning with Limited Sensing Data}

Beyond PGML, there are also other commonly used techniques for addressing the limited data challenge for applying machine and deep learning to sensing data. 
(1) \emph{Ensemble Learning}:
Algorithms like Gradient Boosting \cite{friedman2001greedy} and AdaBoost \cite{freund1995desicion} are effective for small datasets as they combine weak learners to improve accuracy while mitigating overfitting. Ensemble methods also work well in noisy environments by leveraging diverse models.
(2) \emph{Tensor-Based Frameworks}:
FLMS \cite{mullick2024framework} combines user-agnostic and personalized modeling approaches with tensor-based ranking strategies. This method is designed to handle small, multimodal, and noisy sensor datasets effectively, particularly in health-related applications. It addresses overfitting by balancing macro-level patterns with user-specific insights.
(3) \emph{Dimensionality Reduction Techniques}: Preprocessing methods like Principal Component Analysis (PCA) and Wavelet Transforms can optimize small datasets by reducing dimensionality while preserving critical information. These transformations are often integrated with machine learning models to enhance performance on limited sensor data \cite{yotov2024data}.
(4) \emph{Transfer Learning and Data Augmentation}: While deep learning models like Convolutional Neural Networks (CNNs) and Long Short-Term Memory networks (LSTMs) are powerful for larger datasets, their performance deteriorates with limited data due to overfitting. However, transfer learning and data augmentation can make them viable options in some cases \cite{um2017data, wang2022sensor, icsguder2023fedopenhar, ramesh2024transfer}.
(5) \emph{Self-supervised learning}: 
When datasets lack labels due to the high cost and time required for expert annotation, which is common in sensor data, self-supervised learning is often used to extract meaningful data embeddings without labels, enhancing performance on downstream tasks \cite{xu2023rebar, yuan2024self}. In the future, we aim to integrate these methods with our PGML-based GNN model to further enhance air quality estimation in regions with limited sensing data.

\section{Discussion}\label{sec:discussion}
This work demonstrates the effectiveness of \name in imputing PM2.5 values for socioeconomically disadvantaged regions with limited monitoring resources. Below, we discuss the approach's limitations and potential directions for future research.

\textbf{Dynamic GNN for Missing Measurements.}
This work used 41 out of 51 sensors, which had no missing data for over an hour within the investigated period for training and testing. However, in real-world scenarios, sensors often experience data loss or interruptions.
Partial temporally available sensing data could enhance the accuracy of spatial modeling.
We plan to explore customizing dynamic GNN \cite{pytorch_geometric} to flexibly add or remove nodes to incorporate these sensors with partial measurements into the air quality model.

\textbf{Mixture Model for Varied Sensor Densities.} 
The model analysis in Section \ref{subsec:model performance} compares our GNN model and baselines at a fine granularity to further illustrate our model limitation. 
For example, we report that \name performs the best when the sensor density exceeds $0.16/\text{mi}^2$, while IDW and GP outperform it at extremely low densities. 
These analyses provide the guideline for further improving the accuracy across all sensor densities. 
In the future, we will develop a mixture model that integrates \name, IDW, and GP, combining them with learnable weights based on sensor density.

\textbf{Heterogeneous Data Sources and Modalities.}
This work utilizes calibrated citizen science sensor data from a single source, SJVAir. However, considering heterogeneity across multiple data sources is crucial for improving prediction accuracy. A single source may introduce systematic bias while integrating additional data can help mitigate this issue and produce more reliable estimates. For instance, regulatory station measurements, though sparse, offer higher reliability than SJVAir data. Additionally, drone-captured imaging data can identify local pollutant sources, such as industrial plants, which can further enhance the Local Module's ability to model pollutant generation and removal information (Section \ref{subsec:local}). 
In the future, we aim to explore new model architectures and data fusion techniques to effectively and efficiently integrate heterogeneous sources and multiple data modalities for improving prediction accuracy.

\textbf{Large Language Models for Air Quality Analysis.} 
Large language models (LLMs) have been shown to perform human-like reasoning \cite{wei2022chain, yao2023tree, yao2023react, huang2022towards, hu2024large}, planning \cite{huang2024understanding, wei2025plangenllms, aghzal2025survey}, and tool use \cite{qin2023toolllm, qu2025tool} by leveraging their vast real-world knowledge. 
As a result, they have recently been applied to understanding graph \cite{jin2024large, perozzi2024let}  and time series data \cite{zhang2024large, gruver2023large}. 
In the future, we will use LLMs to conduct multifaceted analyses with air quality data, such as air quality prediction, air quality sensor placement planning, and air pollutant source localization reasoning.

\section{Conclusion}\label{sec:conclusion}

In this paper, we introduce \name, a graph-based physics-guided learning scheme tailored for constrained monitoring data.
We leverage the flexibility of Graph Neural Networks in modeling both sensors' data and sensors' relationships to allow more efficient spatial modeling with limited spatial measurements.
To compensate for limited meteorology data, we customize wind-based edge features and a residual message-passing GNN model to make \name more adaptive to coarse wind measurements in monitoring-constrained regions.
We also integrate the physics knowledge from the Convection-Diffusion Equation into our graph learning framework to enhance the model accuracy under high spatial heterogeneity.
We conduct multiple experiments with real-world data from monitoring-constrained regions.
The evaluation shows that \name achieves the highest accuracy and reduces the interpolation errors by 9\% to 56\% compared to multiple baseline models.

\section*{acknowledgement}
This research is supported by a UC Merced Spring 2023 Climate Action Seed Competition Grant, CAHSI-Google Institutional Research Program Award, and F3 R\&D GSR Award funded by the US Department of Commerce, Economic Development Administration Build Back Better Regional Challenge.

\bibliographystyle{ACM-Reference-Format}
\bibliography{acmart}


\begin{thebibliography}{82}


\ifx \showCODEN    \undefined \def \showCODEN     #1{\unskip}     \fi
\ifx \showDOI      \undefined \def \showDOI       #1{#1}\fi
\ifx \showISBNx    \undefined \def \showISBNx     #1{\unskip}     \fi
\ifx \showISBNxiii \undefined \def \showISBNxiii  #1{\unskip}     \fi
\ifx \showISSN     \undefined \def \showISSN      #1{\unskip}     \fi
\ifx \showLCCN     \undefined \def \showLCCN      #1{\unskip}     \fi
\ifx \shownote     \undefined \def \shownote      #1{#1}          \fi
\ifx \showarticletitle \undefined \def \showarticletitle #1{#1}   \fi
\ifx \showURL      \undefined \def \showURL       {\relax}        \fi
\providecommand\bibfield[2]{#2}
\providecommand\bibinfo[2]{#2}
\providecommand\natexlab[1]{#1}
\providecommand\showeprint[2][]{arXiv:#2}

\bibitem[Agency(2024)]%
        {EPA2024}
\bibfield{author}{\bibinfo{person}{U.S. Environmental~Protection Agency}.} \bibinfo{year}{2024}\natexlab{}.
\newblock \bibinfo{title}{Links Between Air Pollution and Childhood Asthma}.
\newblock
\newblock
\urldef\tempurl%
\url{https://www.epa.gov/sciencematters/links-between-air-pollution-and-childhood-asthma}
\showURL{%
\tempurl}
\newblock
\shownote{Accessed: 2024-06-10}.


\bibitem[Aghzal et~al\mbox{.}(2025)]%
        {aghzal2025survey}
\bibfield{author}{\bibinfo{person}{Mohamed Aghzal}, \bibinfo{person}{Erion Plaku}, \bibinfo{person}{Gregory~J Stein}, {and} \bibinfo{person}{Ziyu Yao}.} \bibinfo{year}{2025}\natexlab{}.
\newblock \showarticletitle{A Survey on Large Language Models for Automated Planning}.
\newblock \bibinfo{journal}{\emph{arXiv preprint arXiv:2502.12435}} (\bibinfo{year}{2025}).
\newblock


\bibitem[Air(2024)]%
        {purpleair2024documentation}
\bibfield{author}{\bibinfo{person}{Purple Air}.} \bibinfo{year}{2024}\natexlab{}.
\newblock \bibinfo{title}{Research Considerations}.
\newblock \bibinfo{howpublished}{\url{https://community.purpleair.com/t/research-considerations/9144}}.
\newblock
\newblock
\shownote{Accessed: 2024-07-21}.


\bibitem[Bronstein et~al\mbox{.}(2017)]%
        {bronstein2017geometric}
\bibfield{author}{\bibinfo{person}{Michael~M Bronstein}, \bibinfo{person}{Joan Bruna}, \bibinfo{person}{Yann LeCun}, \bibinfo{person}{Arthur Szlam}, {and} \bibinfo{person}{Pierre Vandergheynst}.} \bibinfo{year}{2017}\natexlab{}.
\newblock \showarticletitle{Geometric deep learning: going beyond euclidean data}.
\newblock \bibinfo{journal}{\emph{IEEE Signal Processing Magazine}} \bibinfo{volume}{34}, \bibinfo{number}{4} (\bibinfo{year}{2017}), \bibinfo{pages}{18--42}.
\newblock


\bibitem[Chen et~al\mbox{.}(2020)]%
        {chen2020adaptive}
\bibfield{author}{\bibinfo{person}{Xinlei Chen}, \bibinfo{person}{Susu Xu}, \bibinfo{person}{Xinyu Liu}, \bibinfo{person}{Xiangxiang Xu}, \bibinfo{person}{Hae~Young Noh}, \bibinfo{person}{Lin Zhang}, {and} \bibinfo{person}{Pei Zhang}.} \bibinfo{year}{2020}\natexlab{}.
\newblock \showarticletitle{Adaptive hybrid model-enabled sensing system (HMSS) for mobile fine-grained air pollution estimation}.
\newblock \bibinfo{journal}{\emph{IEEE Transactions on Mobile Computing}} \bibinfo{volume}{21}, \bibinfo{number}{6} (\bibinfo{year}{2020}), \bibinfo{pages}{1927--1944}.
\newblock


\bibitem[Cheng et~al\mbox{.}(2014)]%
        {cheng2014aircloud}
\bibfield{author}{\bibinfo{person}{Yun Cheng}, \bibinfo{person}{Xiucheng Li}, \bibinfo{person}{Zhijun Li}, \bibinfo{person}{Shouxu Jiang}, \bibinfo{person}{Yilong Li}, \bibinfo{person}{Ji Jia}, {and} \bibinfo{person}{Xiaofan Jiang}.} \bibinfo{year}{2014}\natexlab{}.
\newblock \showarticletitle{AirCloud: A cloud-based air-quality monitoring system for everyone}. In \bibinfo{booktitle}{\emph{Proceedings of the 12th ACM Conference on Embedded Network Sensor Systems}}. \bibinfo{pages}{251--265}.
\newblock


\bibitem[Chow(2024)]%
        {NBCNews2024}
\bibfield{author}{\bibinfo{person}{Denise Chow}.} \bibinfo{year}{2024}\natexlab{}.
\newblock \bibinfo{title}{131 Million in U.S. Live in Areas with Unhealthy Pollution Levels, Lung Association Reports}.
\newblock
\newblock
\urldef\tempurl%
\url{https://www.nbcnews.com/science/environment/131-million-us-live-areas-unhealthy-pollution-levels-lung-association-rcna148795}
\showURL{%
\tempurl}
\newblock
\shownote{Accessed: 2024-06-10}.


\bibitem[Daw et~al\mbox{.}(2022)]%
        {daw2022physics}
\bibfield{author}{\bibinfo{person}{Arka Daw}, \bibinfo{person}{Anuj Karpatne}, \bibinfo{person}{William~D Watkins}, \bibinfo{person}{Jordan~S Read}, {and} \bibinfo{person}{Vipin Kumar}.} \bibinfo{year}{2022}\natexlab{}.
\newblock \showarticletitle{Physics-guided neural networks (pgnn): An application in lake temperature modeling}.
\newblock In \bibinfo{booktitle}{\emph{Knowledge guided machine learning}}. \bibinfo{publisher}{Chapman and Hall/CRC}, \bibinfo{pages}{353--372}.
\newblock


\bibitem[Devarakonda et~al\mbox{.}(2013)]%
        {devarakonda2013real}
\bibfield{author}{\bibinfo{person}{Srinivas Devarakonda}, \bibinfo{person}{Parveen Sevusu}, \bibinfo{person}{Hongzhang Liu}, \bibinfo{person}{Ruilin Liu}, \bibinfo{person}{Liviu Iftode}, {and} \bibinfo{person}{Badri Nath}.} \bibinfo{year}{2013}\natexlab{}.
\newblock \showarticletitle{Real-time air quality monitoring through mobile sensing in metropolitan areas}. In \bibinfo{booktitle}{\emph{Proceedings of the 2nd ACM SIGKDD international workshop on urban computing}}. \bibinfo{pages}{1--8}.
\newblock


\bibitem[District(2021)]%
        {SanJoaquinValleyAirPollutionControlDistrict2021}
\bibfield{author}{\bibinfo{person}{San Joaquin Valley Air Pollution~Control District}.} \bibinfo{year}{2021}\natexlab{}.
\newblock \bibinfo{title}{Air Quality in San Joaquin Valley}.
\newblock
\newblock
\urldef\tempurl%
\url{https://ww2.valleyair.org/media/4x5ng03o/short-presentation-for-web-2021.pdf}
\showURL{%
\tempurl}


\bibitem[Egan and Mahoney(1972)]%
        {egan1972numerical}
\bibfield{author}{\bibinfo{person}{Bruce~A Egan} {and} \bibinfo{person}{James~R Mahoney}.} \bibinfo{year}{1972}\natexlab{}.
\newblock \showarticletitle{Numerical modeling of advection and diffusion of urban area source pollutants}.
\newblock \bibinfo{journal}{\emph{Journal of Applied Meteorology and Climatology}} \bibinfo{volume}{11}, \bibinfo{number}{2} (\bibinfo{year}{1972}), \bibinfo{pages}{312--322}.
\newblock


\bibitem[Engel-Cox et~al\mbox{.}(2004)]%
        {engel2004recommendations}
\bibfield{author}{\bibinfo{person}{Jill~A Engel-Cox}, \bibinfo{person}{Raymond~M Hoff}, {and} \bibinfo{person}{ADJ Haymet}.} \bibinfo{year}{2004}\natexlab{}.
\newblock \showarticletitle{Recommendations on the use of satellite remote-sensing data for urban air quality}.
\newblock \bibinfo{journal}{\emph{Journal of the Air \& Waste Management Association}} \bibinfo{volume}{54}, \bibinfo{number}{11} (\bibinfo{year}{2004}), \bibinfo{pages}{1360--1371}.
\newblock


\bibitem[Ermak(1990)]%
        {ermak1990user}
\bibfield{author}{\bibinfo{person}{Donald~L Ermak}.} \bibinfo{year}{1990}\natexlab{}.
\newblock \bibinfo{booktitle}{\emph{User's manual for SLAB: An atmospheric dispersion model for denser-than-air-releases}}.
\newblock \bibinfo{type}{{T}echnical {R}eport}. \bibinfo{institution}{Lawrence Livermore National Lab., CA (USA)}.
\newblock


\bibitem[Favez et~al\mbox{.}(2010)]%
        {favez2010inter}
\bibfield{author}{\bibinfo{person}{O Favez}, \bibinfo{person}{I El~Haddad}, \bibinfo{person}{C Piot}, \bibinfo{person}{A Bor{\'e}ave}, \bibinfo{person}{E Abidi}, \bibinfo{person}{Nicolas Marchand}, \bibinfo{person}{J-L Jaffrezo}, \bibinfo{person}{J-L Besombes}, \bibinfo{person}{M-B Personnaz}, \bibinfo{person}{J Sciare}, {et~al\mbox{.}}} \bibinfo{year}{2010}\natexlab{}.
\newblock \showarticletitle{Inter-comparison of source apportionment models for the estimation of wood burning aerosols during wintertime in an Alpine city (Grenoble, France)}.
\newblock \bibinfo{journal}{\emph{Atmospheric Chemistry and Physics}} \bibinfo{volume}{10}, \bibinfo{number}{12} (\bibinfo{year}{2010}), \bibinfo{pages}{5295--5314}.
\newblock


\bibitem[Freund and Schapire(1995)]%
        {freund1995desicion}
\bibfield{author}{\bibinfo{person}{Yoav Freund} {and} \bibinfo{person}{Robert~E Schapire}.} \bibinfo{year}{1995}\natexlab{}.
\newblock \showarticletitle{A desicion-theoretic generalization of on-line learning and an application to boosting}. In \bibinfo{booktitle}{\emph{European conference on computational learning theory}}. Springer, \bibinfo{pages}{23--37}.
\newblock


\bibitem[Friedman(2001)]%
        {friedman2001greedy}
\bibfield{author}{\bibinfo{person}{Jerome~H Friedman}.} \bibinfo{year}{2001}\natexlab{}.
\newblock \showarticletitle{Greedy function approximation: a gradient boosting machine}.
\newblock \bibinfo{journal}{\emph{Annals of statistics}} (\bibinfo{year}{2001}), \bibinfo{pages}{1189--1232}.
\newblock


\bibitem[Friedman and Plumer(2024)]%
        {NYTimes2024}
\bibfield{author}{\bibinfo{person}{Lisa Friedman} {and} \bibinfo{person}{Brad Plumer}.} \bibinfo{year}{2024}\natexlab{}.
\newblock \showarticletitle{Biden Administration Announces New Pollution Standards for Power Plants}.
\newblock \bibinfo{journal}{\emph{The New York Times}} (\bibinfo{year}{2024}).
\newblock
\urldef\tempurl%
\url{https://www.nytimes.com/2024/04/25/climate/biden-power-plants-pollution.html}
\showURL{%
\tempurl}
\newblock
\shownote{Accessed: 2024-06-10}.


\bibitem[Gilmer et~al\mbox{.}(2017)]%
        {gilmer2017neural}
\bibfield{author}{\bibinfo{person}{Justin Gilmer}, \bibinfo{person}{Samuel~S Schoenholz}, \bibinfo{person}{Patrick~F Riley}, \bibinfo{person}{Oriol Vinyals}, {and} \bibinfo{person}{George~E Dahl}.} \bibinfo{year}{2017}\natexlab{}.
\newblock \showarticletitle{Neural message passing for quantum chemistry}. In \bibinfo{booktitle}{\emph{International conference on machine learning}}. PMLR, \bibinfo{pages}{1263--1272}.
\newblock


\bibitem[Gobeli et~al\mbox{.}(2008)]%
        {gobeli2008met}
\bibfield{author}{\bibinfo{person}{David Gobeli}, \bibinfo{person}{Herbert Schloesser}, {and} \bibinfo{person}{Thomas Pottberg}.} \bibinfo{year}{2008}\natexlab{}.
\newblock \showarticletitle{Met one instruments BAM-1020 beta attenuation mass monitor US-EPA PM2. 5 federal equivalent method field test results}. In \bibinfo{booktitle}{\emph{The Air \& Waste Management Association (A\&WMA) Conference, Kansas City, MO}}, Vol.~\bibinfo{volume}{2}. Citeseer.
\newblock


\bibitem[Gori et~al\mbox{.}(2005)]%
        {gori2005new}
\bibfield{author}{\bibinfo{person}{Marco Gori}, \bibinfo{person}{Gabriele Monfardini}, {and} \bibinfo{person}{Franco Scarselli}.} \bibinfo{year}{2005}\natexlab{}.
\newblock \showarticletitle{A new model for learning in graph domains}. In \bibinfo{booktitle}{\emph{Proceedings. 2005 IEEE international joint conference on neural networks, 2005.}}, Vol.~\bibinfo{volume}{2}. IEEE, \bibinfo{pages}{729--734}.
\newblock


\bibitem[GridInfo(2024)]%
        {GridInfo_Fresno}
\bibfield{author}{\bibinfo{person}{GridInfo}.} \bibinfo{year}{2024}\natexlab{}.
\newblock \bibinfo{title}{Fresno, CA Electricity Generation Summary}.
\newblock
\newblock
\urldef\tempurl%
\url{https://www.gridinfo.com/california/fresno}
\showURL{%
\tempurl}


\bibitem[Gruver et~al\mbox{.}(2023)]%
        {gruver2023large}
\bibfield{author}{\bibinfo{person}{Nate Gruver}, \bibinfo{person}{Marc Finzi}, \bibinfo{person}{Shikai Qiu}, {and} \bibinfo{person}{Andrew~G Wilson}.} \bibinfo{year}{2023}\natexlab{}.
\newblock \showarticletitle{Large language models are zero-shot time series forecasters}.
\newblock \bibinfo{journal}{\emph{Advances in Neural Information Processing Systems}}  \bibinfo{volume}{36} (\bibinfo{year}{2023}), \bibinfo{pages}{19622--19635}.
\newblock


\bibitem[Hettige et~al\mbox{.}(2024)]%
        {hettige2024airphynet}
\bibfield{author}{\bibinfo{person}{Kethmi~Hirushini Hettige}, \bibinfo{person}{Jiahao Ji}, \bibinfo{person}{Shili Xiang}, \bibinfo{person}{Cheng Long}, \bibinfo{person}{Gao Cong}, {and} \bibinfo{person}{Jingyuan Wang}.} \bibinfo{year}{2024}\natexlab{}.
\newblock \showarticletitle{AirPhyNet: Harnessing Physics-Guided Neural Networks for Air Quality Prediction}.
\newblock \bibinfo{journal}{\emph{arXiv preprint arXiv:2402.03784}} (\bibinfo{year}{2024}).
\newblock


\bibitem[Hsieh et~al\mbox{.}(2015)]%
        {hsieh2015inferring}
\bibfield{author}{\bibinfo{person}{Hsun-Ping Hsieh}, \bibinfo{person}{Shou-De Lin}, {and} \bibinfo{person}{Yu Zheng}.} \bibinfo{year}{2015}\natexlab{}.
\newblock \showarticletitle{Inferring air quality for station location recommendation based on urban big data}. In \bibinfo{booktitle}{\emph{Proceedings of the 21th ACM SIGKDD international conference on knowledge discovery and data mining}}. \bibinfo{pages}{437--446}.
\newblock


\bibitem[Hu et~al\mbox{.}(2023)]%
        {hu2023enhancing}
\bibfield{author}{\bibinfo{person}{Zhizhang Hu}, \bibinfo{person}{Shangjie Du}, \bibinfo{person}{Yuning Chen}, \bibinfo{person}{Xuan Zhang}, \bibinfo{person}{Wan Du}, \bibinfo{person}{Asa Bradman}, {and} \bibinfo{person}{Shijia Pan}.} \bibinfo{year}{2023}\natexlab{}.
\newblock \showarticletitle{Enhancing Fault Resilience of Air Quality Monitoring in San Joaquin Valley: A Data Equity Analysis}. In \bibinfo{booktitle}{\emph{Proceedings of the 21st ACM Conference on Embedded Networked Sensor Systems}}. \bibinfo{pages}{514--515}.
\newblock


\bibitem[Hu et~al\mbox{.}(2022a)]%
        {hu2022ciphy}
\bibfield{author}{\bibinfo{person}{Zhizhang Hu}, \bibinfo{person}{Tong Yu}, \bibinfo{person}{Ruiyi Zhang}, {and} \bibinfo{person}{Shijia Pan}.} \bibinfo{year}{2022}\natexlab{a}.
\newblock \showarticletitle{CIPhy: Causal Intervention with Physical Confounder from IoT Sensor Data for Robust Occupant Information Inference}. In \bibinfo{booktitle}{\emph{Proceedings of the 20th ACM Conference on Embedded Networked Sensor Systems}}. \bibinfo{pages}{966--972}.
\newblock


\bibitem[Hu et~al\mbox{.}(2024)]%
        {hu2024large}
\bibfield{author}{\bibinfo{person}{Zhizhang Hu}, \bibinfo{person}{Yue Zhang}, \bibinfo{person}{Ryan Rossi}, \bibinfo{person}{Tong Yu}, \bibinfo{person}{Sungchul Kim}, {and} \bibinfo{person}{Shijia Pan}.} \bibinfo{year}{2024}\natexlab{}.
\newblock \showarticletitle{Are large language models capable of causal reasoning for sensing data analysis?}. In \bibinfo{booktitle}{\emph{Proceedings of the Workshop on Edge and Mobile Foundation Models}}. \bibinfo{pages}{24--29}.
\newblock


\bibitem[Hu et~al\mbox{.}(2022b)]%
        {hu2022vma}
\bibfield{author}{\bibinfo{person}{Zhizhang Hu}, \bibinfo{person}{Yue Zhang}, \bibinfo{person}{Tong Yu}, {and} \bibinfo{person}{Shijia Pan}.} \bibinfo{year}{2022}\natexlab{b}.
\newblock \showarticletitle{VMA: Domain variance-and modality-aware model transfer for fine-grained occupant activity recognition}. In \bibinfo{booktitle}{\emph{2022 21st ACM/IEEE International Conference on Information Processing in Sensor Networks (IPSN)}}. IEEE, \bibinfo{pages}{259--270}.
\newblock


\bibitem[Huang and Chang(2022)]%
        {huang2022towards}
\bibfield{author}{\bibinfo{person}{Jie Huang} {and} \bibinfo{person}{Kevin Chen-Chuan Chang}.} \bibinfo{year}{2022}\natexlab{}.
\newblock \showarticletitle{Towards reasoning in large language models: A survey}.
\newblock \bibinfo{journal}{\emph{arXiv preprint arXiv:2212.10403}} (\bibinfo{year}{2022}).
\newblock


\bibitem[Huang et~al\mbox{.}(2024)]%
        {huang2024understanding}
\bibfield{author}{\bibinfo{person}{Xu Huang}, \bibinfo{person}{Weiwen Liu}, \bibinfo{person}{Xiaolong Chen}, \bibinfo{person}{Xingmei Wang}, \bibinfo{person}{Hao Wang}, \bibinfo{person}{Defu Lian}, \bibinfo{person}{Yasheng Wang}, \bibinfo{person}{Ruiming Tang}, {and} \bibinfo{person}{Enhong Chen}.} \bibinfo{year}{2024}\natexlab{}.
\newblock \showarticletitle{Understanding the planning of LLM agents: A survey}.
\newblock \bibinfo{journal}{\emph{arXiv preprint arXiv:2402.02716}} (\bibinfo{year}{2024}).
\newblock


\bibitem[{\.I}{\c{s}}g{\"u}der and {\.I}ncel(2023)]%
        {icsguder2023fedopenhar}
\bibfield{author}{\bibinfo{person}{Egemen {\.I}{\c{s}}g{\"u}der} {and} \bibinfo{person}{{\"O}zlem~Durmaz {\.I}ncel}.} \bibinfo{year}{2023}\natexlab{}.
\newblock \showarticletitle{FedOpenHAR: Federated Multi-Task Transfer Learning for Sensor-Based Human Activity Recognition}.
\newblock \bibinfo{journal}{\emph{arXiv preprint arXiv:2311.07765}} (\bibinfo{year}{2023}).
\newblock


\bibitem[Janssen et~al\mbox{.}(2008)]%
        {janssen2008spatial}
\bibfield{author}{\bibinfo{person}{Stijn Janssen}, \bibinfo{person}{Gerwin Dumont}, \bibinfo{person}{Frans Fierens}, {and} \bibinfo{person}{Clemens Mensink}.} \bibinfo{year}{2008}\natexlab{}.
\newblock \showarticletitle{Spatial interpolation of air pollution measurements using CORINE land cover data}.
\newblock \bibinfo{journal}{\emph{Atmospheric Environment}} \bibinfo{volume}{42}, \bibinfo{number}{20} (\bibinfo{year}{2008}), \bibinfo{pages}{4884--4903}.
\newblock


\bibitem[Jin et~al\mbox{.}(2024)]%
        {jin2024large}
\bibfield{author}{\bibinfo{person}{Bowen Jin}, \bibinfo{person}{Gang Liu}, \bibinfo{person}{Chi Han}, \bibinfo{person}{Meng Jiang}, \bibinfo{person}{Heng Ji}, {and} \bibinfo{person}{Jiawei Han}.} \bibinfo{year}{2024}\natexlab{}.
\newblock \showarticletitle{Large language models on graphs: A comprehensive survey}.
\newblock \bibinfo{journal}{\emph{IEEE Transactions on Knowledge and Data Engineering}} (\bibinfo{year}{2024}).
\newblock


\bibitem[Kaimaoge(2024)]%
        {ignnk2024}
\bibfield{author}{\bibinfo{person}{Kaimaoge}.} \bibinfo{year}{2024}\natexlab{}.
\newblock \bibinfo{title}{IGNNK: Inductive Graph Neural Networks for Spatiotemporal Kriging}.
\newblock \bibinfo{howpublished}{\url{https://github.com/Kaimaoge/IGNNK}}.
\newblock


\bibitem[Kelly and Fussell(2015)]%
        {kelly2015air}
\bibfield{author}{\bibinfo{person}{Frank~J Kelly} {and} \bibinfo{person}{Julia~C Fussell}.} \bibinfo{year}{2015}\natexlab{}.
\newblock \showarticletitle{Air pollution and public health: emerging hazards and improved understanding of risk}.
\newblock \bibinfo{journal}{\emph{Environmental geochemistry and health}}  \bibinfo{volume}{37} (\bibinfo{year}{2015}), \bibinfo{pages}{631--649}.
\newblock


\bibitem[Kipf and Welling(2016)]%
        {kipf2016semi}
\bibfield{author}{\bibinfo{person}{Thomas~N Kipf} {and} \bibinfo{person}{Max Welling}.} \bibinfo{year}{2016}\natexlab{}.
\newblock \showarticletitle{Semi-supervised classification with graph convolutional networks}.
\newblock \bibinfo{journal}{\emph{arXiv preprint arXiv:1609.02907}} (\bibinfo{year}{2016}).
\newblock


\bibitem[Liu et~al\mbox{.}(2022)]%
        {liu2022statistical}
\bibfield{author}{\bibinfo{person}{Xiao Liu}, \bibinfo{person}{Kyongmin Yeo}, {and} \bibinfo{person}{Siyuan Lu}.} \bibinfo{year}{2022}\natexlab{}.
\newblock \showarticletitle{Statistical modeling for spatio-temporal data from stochastic convection-diffusion processes}.
\newblock \bibinfo{journal}{\emph{J. Amer. Statist. Assoc.}} \bibinfo{volume}{117}, \bibinfo{number}{539} (\bibinfo{year}{2022}), \bibinfo{pages}{1482--1499}.
\newblock


\bibitem[Liu et~al\mbox{.}(2024)]%
        {liu2024mobiair}
\bibfield{author}{\bibinfo{person}{Yuxuan Liu}, \bibinfo{person}{Haoyang Wang}, \bibinfo{person}{Fanhang Man}, \bibinfo{person}{Jingao Xu}, \bibinfo{person}{Fan Dang}, \bibinfo{person}{Yunhao Liu}, \bibinfo{person}{Xiao-Ping Zhang}, {and} \bibinfo{person}{Xinlei Chen}.} \bibinfo{year}{2024}\natexlab{}.
\newblock \showarticletitle{MobiAir: Unleashing Sensor Mobility for City-scale and Fine-grained Air-Quality Monitoring with AirBERT}. In \bibinfo{booktitle}{\emph{Proceedings of the 22nd Annual International Conference on Mobile Systems, Applications and Services}}. \bibinfo{pages}{223--236}.
\newblock


\bibitem[Ma et~al\mbox{.}(2019)]%
        {ma2019temporal}
\bibfield{author}{\bibinfo{person}{Jun Ma}, \bibinfo{person}{Yuexiong Ding}, \bibinfo{person}{Jack~CP Cheng}, \bibinfo{person}{Feifeng Jiang}, {and} \bibinfo{person}{Zhiwei Wan}.} \bibinfo{year}{2019}\natexlab{}.
\newblock \showarticletitle{A temporal-spatial interpolation and extrapolation method based on geographic Long Short-Term Memory neural network for PM2. 5}.
\newblock \bibinfo{journal}{\emph{Journal of Cleaner Production}}  \bibinfo{volume}{237} (\bibinfo{year}{2019}), \bibinfo{pages}{117729}.
\newblock


\bibitem[Martin(2008)]%
        {martin2008satellite}
\bibfield{author}{\bibinfo{person}{Randall~V Martin}.} \bibinfo{year}{2008}\natexlab{}.
\newblock \showarticletitle{Satellite remote sensing of surface air quality}.
\newblock \bibinfo{journal}{\emph{Atmospheric environment}} \bibinfo{volume}{42}, \bibinfo{number}{34} (\bibinfo{year}{2008}), \bibinfo{pages}{7823--7843}.
\newblock


\bibitem[Mirshekari et~al\mbox{.}(2020)]%
        {mirshekari2020step}
\bibfield{author}{\bibinfo{person}{Mostafa Mirshekari}, \bibinfo{person}{Jonathon Fagert}, \bibinfo{person}{Shijia Pan}, \bibinfo{person}{Pei Zhang}, {and} \bibinfo{person}{Hae~Young Noh}.} \bibinfo{year}{2020}\natexlab{}.
\newblock \showarticletitle{Step-level occupant detection across different structures through footstep-induced floor vibration using model transfer}.
\newblock \bibinfo{journal}{\emph{Journal of Engineering Mechanics}} \bibinfo{volume}{146}, \bibinfo{number}{3} (\bibinfo{year}{2020}), \bibinfo{pages}{04019137}.
\newblock


\bibitem[Mullick et~al\mbox{.}(2024)]%
        {mullick2024framework}
\bibfield{author}{\bibinfo{person}{Tahsin Mullick}, \bibinfo{person}{Sam Shaaban}, \bibinfo{person}{Ana Radovic}, \bibinfo{person}{Afsaneh Doryab}, {et~al\mbox{.}}} \bibinfo{year}{2024}\natexlab{}.
\newblock \showarticletitle{Framework for ranking machine learning predictions of limited, multimodal, and longitudinal behavioral passive sensing data: combining user-agnostic and personalized modeling}.
\newblock \bibinfo{journal}{\emph{JMIR AI}} \bibinfo{volume}{3}, \bibinfo{number}{1} (\bibinfo{year}{2024}), \bibinfo{pages}{e47805}.
\newblock


\bibitem[Patel et~al\mbox{.}(2022)]%
        {patel2022accurate}
\bibfield{author}{\bibinfo{person}{Zeel~B Patel}, \bibinfo{person}{Palak Purohit}, \bibinfo{person}{Harsh~M Patel}, \bibinfo{person}{Shivam Sahni}, {and} \bibinfo{person}{Nipun Batra}.} \bibinfo{year}{2022}\natexlab{}.
\newblock \showarticletitle{Accurate and scalable gaussian processes for fine-grained air quality inference}. In \bibinfo{booktitle}{\emph{Proceedings of the AAAI Conference on Artificial Intelligence}}, Vol.~\bibinfo{volume}{36}. \bibinfo{pages}{12080--12088}.
\newblock


\bibitem[Perozzi et~al\mbox{.}(2024)]%
        {perozzi2024let}
\bibfield{author}{\bibinfo{person}{Bryan Perozzi}, \bibinfo{person}{Bahare Fatemi}, \bibinfo{person}{Dustin Zelle}, \bibinfo{person}{Anton Tsitsulin}, \bibinfo{person}{Mehran Kazemi}, \bibinfo{person}{Rami Al-Rfou}, {and} \bibinfo{person}{Jonathan Halcrow}.} \bibinfo{year}{2024}\natexlab{}.
\newblock \showarticletitle{Let your graph do the talking: Encoding structured data for llms}.
\newblock \bibinfo{journal}{\emph{arXiv preprint arXiv:2402.05862}} (\bibinfo{year}{2024}).
\newblock


\bibitem[Pielke and Uliasz(1998)]%
        {pielke1998use}
\bibfield{author}{\bibinfo{person}{Roger~A Pielke} {and} \bibinfo{person}{M Uliasz}.} \bibinfo{year}{1998}\natexlab{}.
\newblock \showarticletitle{Use of meteorological models as input to regional and mesoscale air quality models—limitations and strengths}.
\newblock \bibinfo{journal}{\emph{Atmospheric environment}} \bibinfo{volume}{32}, \bibinfo{number}{8} (\bibinfo{year}{1998}), \bibinfo{pages}{1455--1466}.
\newblock


\bibitem[Pinedo and Linn(2024)]%
        {pinedo2024fresno}
\bibfield{author}{\bibinfo{person}{Jacqueline Pinedo} {and} \bibinfo{person}{Sarah Linn}.} \bibinfo{year}{2024}\natexlab{}.
\newblock \showarticletitle{Fresno is among top 10 most polluted cities in U.S., American Lung Association. Here’s why}.
\newblock \bibinfo{journal}{\emph{The Fresno Bee}} (\bibinfo{year}{2024}).
\newblock
\urldef\tempurl%
\url{https://www.fresnobee.com/news/local/article288061190.html}
\showURL{%
\tempurl}


\bibitem[{PurpleAir}(2024)]%
        {purpleair}
\bibfield{author}{\bibinfo{person}{{PurpleAir}}.} \bibinfo{year}{2024}\natexlab{}.
\newblock \bibinfo{title}{PurpleAir: Monitoring Air Quality}.
\newblock \bibinfo{howpublished}{\url{https://www2.purpleair.com/}}.
\newblock
\newblock
\shownote{Accessed: 2024-07-23}.


\bibitem[{PyTorch Geometric}(2024)]%
        {pytorch_geometric}
\bibfield{author}{\bibinfo{person}{{PyTorch Geometric}}.} \bibinfo{year}{2024}\natexlab{}.
\newblock \bibinfo{title}{PyTorch Geometric Documentation: Introduction-Minibatches}.
\newblock \bibinfo{howpublished}{\url{https://pytorch-geometric.readthedocs.io/en/latest/get_started/introduction.html}}.
\newblock
\newblock
\shownote{Accessed: 2024-07-23}.


\bibitem[Qin et~al\mbox{.}(2023)]%
        {qin2023toolllm}
\bibfield{author}{\bibinfo{person}{Yujia Qin}, \bibinfo{person}{Shihao Liang}, \bibinfo{person}{Yining Ye}, \bibinfo{person}{Kunlun Zhu}, \bibinfo{person}{Lan Yan}, \bibinfo{person}{Yaxi Lu}, \bibinfo{person}{Yankai Lin}, \bibinfo{person}{Xin Cong}, \bibinfo{person}{Xiangru Tang}, \bibinfo{person}{Bill Qian}, {et~al\mbox{.}}} \bibinfo{year}{2023}\natexlab{}.
\newblock \showarticletitle{Toolllm: Facilitating large language models to master 16000+ real-world apis}.
\newblock \bibinfo{journal}{\emph{arXiv preprint arXiv:2307.16789}} (\bibinfo{year}{2023}).
\newblock


\bibitem[Qu et~al\mbox{.}(2025)]%
        {qu2025tool}
\bibfield{author}{\bibinfo{person}{Changle Qu}, \bibinfo{person}{Sunhao Dai}, \bibinfo{person}{Xiaochi Wei}, \bibinfo{person}{Hengyi Cai}, \bibinfo{person}{Shuaiqiang Wang}, \bibinfo{person}{Dawei Yin}, \bibinfo{person}{Jun Xu}, {and} \bibinfo{person}{Ji-Rong Wen}.} \bibinfo{year}{2025}\natexlab{}.
\newblock \showarticletitle{Tool learning with large language models: A survey}.
\newblock \bibinfo{journal}{\emph{Frontiers of Computer Science}} \bibinfo{volume}{19}, \bibinfo{number}{8} (\bibinfo{year}{2025}), \bibinfo{pages}{198343}.
\newblock


\bibitem[Ramesh et~al\mbox{.}(2024)]%
        {ramesh2024transfer}
\bibfield{author}{\bibinfo{person}{K Ramesh}, \bibinfo{person}{B Rajarao}, \bibinfo{person}{Hitesh~E Chaudhari}, \bibinfo{person}{S~Angel~Latha Mary}, \bibinfo{person}{M Venkatanaresh}, {and} \bibinfo{person}{Tarun~Kumar Dhiman}.} \bibinfo{year}{2024}\natexlab{}.
\newblock \showarticletitle{Transfer learning approach to reduce similar IOT sensor data for industrial applications}.
\newblock \bibinfo{journal}{\emph{Measurement: Sensors}}  \bibinfo{volume}{31} (\bibinfo{year}{2024}), \bibinfo{pages}{100985}.
\newblock


\bibitem[Randerson(1970)]%
        {randerson1970numerical}
\bibfield{author}{\bibinfo{person}{Darryl Randerson}.} \bibinfo{year}{1970}\natexlab{}.
\newblock \showarticletitle{A numerical experiment in simulating the transport of sulfur dioxide through the atmosphere}.
\newblock \bibinfo{journal}{\emph{Atmospheric Environment (1967)}} \bibinfo{volume}{4}, \bibinfo{number}{6} (\bibinfo{year}{1970}), \bibinfo{pages}{615--632}.
\newblock


\bibitem[Rohal et~al\mbox{.}(2024)]%
        {rohal2024don}
\bibfield{author}{\bibinfo{person}{Shubham Rohal}, \bibinfo{person}{Dong~Yoon Lee}, \bibinfo{person}{Carlos Ruiz}, \bibinfo{person}{Joshua Zhang}, \bibinfo{person}{Jonathon Fagert}, \bibinfo{person}{Jun Han}, {and} \bibinfo{person}{Shijia Pan}.} \bibinfo{year}{2024}\natexlab{}.
\newblock \showarticletitle{Don’t Crosstalk to Me: Origami Structure-Augmented Sensing for Scalable Surface Pressure Monitoring}. In \bibinfo{booktitle}{\emph{Proceedings of the 22nd ACM Conference on Embedded Networked Sensor Systems}}. \bibinfo{pages}{493--506}.
\newblock


\bibitem[Scarselli et~al\mbox{.}(2008)]%
        {scarselli2008graph}
\bibfield{author}{\bibinfo{person}{Franco Scarselli}, \bibinfo{person}{Marco Gori}, \bibinfo{person}{Ah~Chung Tsoi}, \bibinfo{person}{Markus Hagenbuchner}, {and} \bibinfo{person}{Gabriele Monfardini}.} \bibinfo{year}{2008}\natexlab{}.
\newblock \showarticletitle{The graph neural network model}.
\newblock \bibinfo{journal}{\emph{IEEE transactions on neural networks}} \bibinfo{volume}{20}, \bibinfo{number}{1} (\bibinfo{year}{2008}), \bibinfo{pages}{61--80}.
\newblock


\bibitem[Schneider(1992)]%
        {schneider1992introduction}
\bibfield{author}{\bibinfo{person}{Stephen~H Schneider}.} \bibinfo{year}{1992}\natexlab{}.
\newblock \showarticletitle{Introduction to climate modeling}.
\newblock \bibinfo{journal}{\emph{SMR}} \bibinfo{volume}{648}, \bibinfo{number}{6} (\bibinfo{year}{1992}), \bibinfo{pages}{1677}.
\newblock


\bibitem[Scire et~al\mbox{.}(2000)]%
        {scire2000user}
\bibfield{author}{\bibinfo{person}{Joseph~S Scire}, \bibinfo{person}{David~G Strimaitis}, \bibinfo{person}{Robert~J Yamartino}, {et~al\mbox{.}}} \bibinfo{year}{2000}\natexlab{}.
\newblock \showarticletitle{A user’s guide for the CALPUFF dispersion model}.
\newblock \bibinfo{journal}{\emph{Earth Tech, Inc}}  \bibinfo{volume}{521} (\bibinfo{year}{2000}), \bibinfo{pages}{1--521}.
\newblock


\bibitem[Shepard(1968)]%
        {shepard1968two}
\bibfield{author}{\bibinfo{person}{Donald Shepard}.} \bibinfo{year}{1968}\natexlab{}.
\newblock \showarticletitle{A two-dimensional interpolation function for irregularly-spaced data}. In \bibinfo{booktitle}{\emph{Proceedings of the 1968 23rd ACM national conference}}. \bibinfo{pages}{517--524}.
\newblock


\bibitem[Shir and Shieh(1974)]%
        {shir1974generalized}
\bibfield{author}{\bibinfo{person}{CC Shir} {and} \bibinfo{person}{LJ Shieh}.} \bibinfo{year}{1974}\natexlab{}.
\newblock \showarticletitle{A generalized urban air pollution model and its application to the study of SO2 distributions in the St. Louis metropolitan area}.
\newblock \bibinfo{journal}{\emph{Journal of Applied Meteorology and Climatology}} \bibinfo{volume}{13}, \bibinfo{number}{2} (\bibinfo{year}{1974}), \bibinfo{pages}{185--204}.
\newblock


\bibitem[SJVAir(2024)]%
        {sjvair2024website}
\bibfield{author}{\bibinfo{person}{SJVAir}.} \bibinfo{year}{2024}\natexlab{}.
\newblock \bibinfo{title}{SJVAir Official Website}.
\newblock \bibinfo{howpublished}{\url{https://www.sjvair.com/}}.
\newblock
\newblock
\shownote{Accessed: 2024-07-21}.


\bibitem[Stocker(2011)]%
        {stocker2011introduction}
\bibfield{author}{\bibinfo{person}{Thomas Stocker}.} \bibinfo{year}{2011}\natexlab{}.
\newblock \bibinfo{booktitle}{\emph{Introduction to climate modelling}}.
\newblock \bibinfo{publisher}{Springer Science \& Business Media}.
\newblock


\bibitem[Stockie(2011)]%
        {stockie2011mathematics}
\bibfield{author}{\bibinfo{person}{John~M Stockie}.} \bibinfo{year}{2011}\natexlab{}.
\newblock \showarticletitle{The mathematics of atmospheric dispersion modeling}.
\newblock \bibinfo{journal}{\emph{Siam Review}} \bibinfo{volume}{53}, \bibinfo{number}{2} (\bibinfo{year}{2011}), \bibinfo{pages}{349--372}.
\newblock


\bibitem[Um et~al\mbox{.}(2017)]%
        {um2017data}
\bibfield{author}{\bibinfo{person}{Terry~T Um}, \bibinfo{person}{Franz~MJ Pfister}, \bibinfo{person}{Daniel Pichler}, \bibinfo{person}{Satoshi Endo}, \bibinfo{person}{Muriel Lang}, \bibinfo{person}{Sandra Hirche}, \bibinfo{person}{Urban Fietzek}, {and} \bibinfo{person}{Dana Kuli{\'c}}.} \bibinfo{year}{2017}\natexlab{}.
\newblock \showarticletitle{Data augmentation of wearable sensor data for parkinson’s disease monitoring using convolutional neural networks}. In \bibinfo{booktitle}{\emph{Proceedings of the 19th ACM international conference on multimodal interaction}}. \bibinfo{pages}{216--220}.
\newblock


\bibitem[{U.S. Department of Energy}({[n.\,d.]})]%
        {fuel_economy}
\bibfield{author}{\bibinfo{person}{{U.S. Department of Energy}}.} \bibinfo{year}{[n.\,d.]}\natexlab{}.
\newblock \bibinfo{title}{Fuel economy in cold weather}.
\newblock
\newblock
\urldef\tempurl%
\url{https://www.fueleconomy.gov/feg/coldweather.shtml}
\showURL{%
\tempurl}
\newblock
\shownote{Accessed: 2024-07-22}.


\bibitem[{Visual Crossing}(2024)]%
        {visualcrossing}
\bibfield{author}{\bibinfo{person}{{Visual Crossing}}.} \bibinfo{year}{2024}\natexlab{}.
\newblock \bibinfo{title}{Historical Weather Data \& Weather Forecast Data}.
\newblock
\newblock
\urldef\tempurl%
\url{https://www.visualcrossing.com/weather-data}
\showURL{%
\tempurl}


\bibitem[Wackernagel and Wackernagel(2003)]%
        {wackernagel2003ordinary}
\bibfield{author}{\bibinfo{person}{Hans Wackernagel} {and} \bibinfo{person}{Hans Wackernagel}.} \bibinfo{year}{2003}\natexlab{}.
\newblock \showarticletitle{Ordinary kriging}.
\newblock \bibinfo{journal}{\emph{Multivariate geostatistics: An introduction with applications}} (\bibinfo{year}{2003}), \bibinfo{pages}{79--88}.
\newblock


\bibitem[Wang et~al\mbox{.}(2022)]%
        {wang2022sensor}
\bibfield{author}{\bibinfo{person}{Jinqiang Wang}, \bibinfo{person}{Tao Zhu}, \bibinfo{person}{Jingyuan Gan}, \bibinfo{person}{Liming~Luke Chen}, \bibinfo{person}{Huansheng Ning}, {and} \bibinfo{person}{Yaping Wan}.} \bibinfo{year}{2022}\natexlab{}.
\newblock \showarticletitle{Sensor data augmentation by resampling in contrastive learning for human activity recognition}.
\newblock \bibinfo{journal}{\emph{IEEE Sensors Journal}} \bibinfo{volume}{22}, \bibinfo{number}{23} (\bibinfo{year}{2022}), \bibinfo{pages}{22994--23008}.
\newblock


\bibitem[Wei et~al\mbox{.}(2024)]%
        {wei2024temporally}
\bibfield{author}{\bibinfo{person}{Hui Wei}, \bibinfo{person}{Maxwell~A Xu}, \bibinfo{person}{Colin Samplawski}, \bibinfo{person}{James~M Rehg}, \bibinfo{person}{Santosh Kumar}, {and} \bibinfo{person}{Benjamin~M Marlin}.} \bibinfo{year}{2024}\natexlab{}.
\newblock \showarticletitle{Temporally Multi-Scale Sparse Self-Attention for Physical Activity Data Imputation}.
\newblock \bibinfo{journal}{\emph{Proceedings of machine learning research}}  \bibinfo{volume}{248} (\bibinfo{year}{2024}), \bibinfo{pages}{137}.
\newblock


\bibitem[Wei et~al\mbox{.}(2025)]%
        {wei2025plangenllms}
\bibfield{author}{\bibinfo{person}{Hui Wei}, \bibinfo{person}{Zihao Zhang}, \bibinfo{person}{Shenghua He}, \bibinfo{person}{Tian Xia}, \bibinfo{person}{Shijia Pan}, {and} \bibinfo{person}{Fei Liu}.} \bibinfo{year}{2025}\natexlab{}.
\newblock \showarticletitle{PlanGenLLMs: A Modern Survey of LLM Planning Capabilities}.
\newblock \bibinfo{journal}{\emph{arXiv preprint arXiv:2502.11221}} (\bibinfo{year}{2025}).
\newblock


\bibitem[Wei et~al\mbox{.}(2022)]%
        {wei2022chain}
\bibfield{author}{\bibinfo{person}{Jason Wei}, \bibinfo{person}{Xuezhi Wang}, \bibinfo{person}{Dale Schuurmans}, \bibinfo{person}{Maarten Bosma}, \bibinfo{person}{Fei Xia}, \bibinfo{person}{Ed Chi}, \bibinfo{person}{Quoc~V Le}, \bibinfo{person}{Denny Zhou}, {et~al\mbox{.}}} \bibinfo{year}{2022}\natexlab{}.
\newblock \showarticletitle{Chain-of-thought prompting elicits reasoning in large language models}.
\newblock \bibinfo{journal}{\emph{Advances in neural information processing systems}}  \bibinfo{volume}{35} (\bibinfo{year}{2022}), \bibinfo{pages}{24824--24837}.
\newblock


\bibitem[Wu et~al\mbox{.}(2021)]%
        {wu2021inductive}
\bibfield{author}{\bibinfo{person}{Yuankai Wu}, \bibinfo{person}{Dingyi Zhuang}, \bibinfo{person}{Aurelie Labbe}, {and} \bibinfo{person}{Lijun Sun}.} \bibinfo{year}{2021}\natexlab{}.
\newblock \showarticletitle{Inductive graph neural networks for spatiotemporal kriging}. In \bibinfo{booktitle}{\emph{Proceedings of the AAAI Conference on Artificial Intelligence}}, Vol.~\bibinfo{volume}{35}. \bibinfo{pages}{4478--4485}.
\newblock


\bibitem[Wu et~al\mbox{.}(2020)]%
        {wu2020comprehensive}
\bibfield{author}{\bibinfo{person}{Zonghan Wu}, \bibinfo{person}{Shirui Pan}, \bibinfo{person}{Fengwen Chen}, \bibinfo{person}{Guodong Long}, \bibinfo{person}{Chengqi Zhang}, {and} \bibinfo{person}{S~Yu Philip}.} \bibinfo{year}{2020}\natexlab{}.
\newblock \showarticletitle{A comprehensive survey on graph neural networks}.
\newblock \bibinfo{journal}{\emph{IEEE transactions on neural networks and learning systems}} \bibinfo{volume}{32}, \bibinfo{number}{1} (\bibinfo{year}{2020}), \bibinfo{pages}{4--24}.
\newblock


\bibitem[Xu et~al\mbox{.}(2023)]%
        {xu2023rebar}
\bibfield{author}{\bibinfo{person}{Maxwell~A Xu}, \bibinfo{person}{Alexander Moreno}, \bibinfo{person}{Hui Wei}, \bibinfo{person}{Benjamin~M Marlin}, {and} \bibinfo{person}{James~M Rehg}.} \bibinfo{year}{2023}\natexlab{}.
\newblock \showarticletitle{Rebar: Retrieval-based reconstruction for time-series contrastive learning}.
\newblock \bibinfo{journal}{\emph{arXiv preprint arXiv:2311.00519}} (\bibinfo{year}{2023}).
\newblock


\bibitem[Yanosky and Maclntosh(2001)]%
        {yanosky2001comparison}
\bibfield{author}{\bibinfo{person}{Jeff~D Yanosky} {and} \bibinfo{person}{David~L Maclntosh}.} \bibinfo{year}{2001}\natexlab{}.
\newblock \showarticletitle{A comparison of four gravimetric fine particle sampling methods}.
\newblock \bibinfo{journal}{\emph{Journal of the Air \& Waste Management Association}} \bibinfo{volume}{51}, \bibinfo{number}{6} (\bibinfo{year}{2001}), \bibinfo{pages}{878--884}.
\newblock


\bibitem[Yao et~al\mbox{.}(2023a)]%
        {yao2023tree}
\bibfield{author}{\bibinfo{person}{Shunyu Yao}, \bibinfo{person}{Dian Yu}, \bibinfo{person}{Jeffrey Zhao}, \bibinfo{person}{Izhak Shafran}, \bibinfo{person}{Tom Griffiths}, \bibinfo{person}{Yuan Cao}, {and} \bibinfo{person}{Karthik Narasimhan}.} \bibinfo{year}{2023}\natexlab{a}.
\newblock \showarticletitle{Tree of thoughts: Deliberate problem solving with large language models}.
\newblock \bibinfo{journal}{\emph{Advances in neural information processing systems}}  \bibinfo{volume}{36} (\bibinfo{year}{2023}), \bibinfo{pages}{11809--11822}.
\newblock


\bibitem[Yao et~al\mbox{.}(2023b)]%
        {yao2023react}
\bibfield{author}{\bibinfo{person}{Shunyu Yao}, \bibinfo{person}{Jeffrey Zhao}, \bibinfo{person}{Dian Yu}, \bibinfo{person}{Nan Du}, \bibinfo{person}{Izhak Shafran}, \bibinfo{person}{Karthik Narasimhan}, {and} \bibinfo{person}{Yuan Cao}.} \bibinfo{year}{2023}\natexlab{b}.
\newblock \showarticletitle{React: Synergizing reasoning and acting in language models}. In \bibinfo{booktitle}{\emph{International Conference on Learning Representations (ICLR)}}.
\newblock


\bibitem[Ye and Ji(2021)]%
        {ye2021sparse}
\bibfield{author}{\bibinfo{person}{Yang Ye} {and} \bibinfo{person}{Shihao Ji}.} \bibinfo{year}{2021}\natexlab{}.
\newblock \showarticletitle{Sparse graph attention networks}.
\newblock \bibinfo{journal}{\emph{IEEE Transactions on Knowledge and Data Engineering}} \bibinfo{volume}{35}, \bibinfo{number}{1} (\bibinfo{year}{2021}), \bibinfo{pages}{905--916}.
\newblock


\bibitem[Yotov and Aleksieva-Petrova(2024)]%
        {yotov2024data}
\bibfield{author}{\bibinfo{person}{Ognyan Yotov} {and} \bibinfo{person}{Adelina Aleksieva-Petrova}.} \bibinfo{year}{2024}\natexlab{}.
\newblock \showarticletitle{Data-driven prediction model for analysis of sensor data}.
\newblock \bibinfo{journal}{\emph{Electronics}} \bibinfo{volume}{13}, \bibinfo{number}{10} (\bibinfo{year}{2024}), \bibinfo{pages}{1799}.
\newblock


\bibitem[Yu et~al\mbox{.}(2021)]%
        {yu2021vibration}
\bibfield{author}{\bibinfo{person}{Tong Yu}, \bibinfo{person}{Yue Zhang}, \bibinfo{person}{Zhizhang Hu}, \bibinfo{person}{Susu Xu}, {and} \bibinfo{person}{Shijia Pan}.} \bibinfo{year}{2021}\natexlab{}.
\newblock \showarticletitle{Vibration-based indoor human sensing quality reinforcement via thompson sampling}. In \bibinfo{booktitle}{\emph{Proceedings of the First International Workshop on Cyber-Physical-Human System Design and Implementation}}. \bibinfo{pages}{33--38}.
\newblock


\bibitem[Yuan et~al\mbox{.}(2024)]%
        {yuan2024self}
\bibfield{author}{\bibinfo{person}{Hang Yuan}, \bibinfo{person}{Shing Chan}, \bibinfo{person}{Andrew~P Creagh}, \bibinfo{person}{Catherine Tong}, \bibinfo{person}{Aidan Acquah}, \bibinfo{person}{David~A Clifton}, {and} \bibinfo{person}{Aiden Doherty}.} \bibinfo{year}{2024}\natexlab{}.
\newblock \showarticletitle{Self-supervised learning for human activity recognition using 700,000 person-days of wearable data}.
\newblock \bibinfo{journal}{\emph{NPJ digital medicine}} \bibinfo{volume}{7}, \bibinfo{number}{1} (\bibinfo{year}{2024}), \bibinfo{pages}{91}.
\newblock


\bibitem[Zhang et~al\mbox{.}(2020)]%
        {zhang2020physics}
\bibfield{author}{\bibinfo{person}{Ruiyang Zhang}, \bibinfo{person}{Yang Liu}, {and} \bibinfo{person}{Hao Sun}.} \bibinfo{year}{2020}\natexlab{}.
\newblock \showarticletitle{Physics-guided convolutional neural network (PhyCNN) for data-driven seismic response modeling}.
\newblock \bibinfo{journal}{\emph{Engineering Structures}}  \bibinfo{volume}{215} (\bibinfo{year}{2020}), \bibinfo{pages}{110704}.
\newblock


\bibitem[Zhang et~al\mbox{.}(2024)]%
        {zhang2024large}
\bibfield{author}{\bibinfo{person}{Xiyuan Zhang}, \bibinfo{person}{Ranak~Roy Chowdhury}, \bibinfo{person}{Rajesh~K Gupta}, {and} \bibinfo{person}{Jingbo Shang}.} \bibinfo{year}{2024}\natexlab{}.
\newblock \showarticletitle{Large language models for time series: A survey}.
\newblock \bibinfo{journal}{\emph{arXiv preprint arXiv:2402.01801}} (\bibinfo{year}{2024}).
\newblock


\bibitem[Zheng et~al\mbox{.}(2013)]%
        {zheng2013u}
\bibfield{author}{\bibinfo{person}{Yu Zheng}, \bibinfo{person}{Furui Liu}, {and} \bibinfo{person}{Hsun-Ping Hsieh}.} \bibinfo{year}{2013}\natexlab{}.
\newblock \showarticletitle{U-air: When urban air quality inference meets big data}. In \bibinfo{booktitle}{\emph{Proceedings of the 19th ACM SIGKDD international conference on Knowledge discovery and data mining}}. \bibinfo{pages}{1436--1444}.
\newblock


\end{thebibliography}

\end{document}